\documentclass[10pt]{article} %
\usepackage[accepted]{tmlr}

\usepackage{amsmath,amsfonts,bm}

\def\eqref#1{equation~\ref{#1}}

\def\1{\bm{1}}

\def\mO{{\bm{O}}}

\def\mR{{\bm{R}}}

\DeclareMathAlphabet{\mathsfit}{\encodingdefault}{\sfdefault}{m}{sl}
\SetMathAlphabet{\mathsfit}{bold}{\encodingdefault}{\sfdefault}{bx}{n}

\usepackage{hyperref}
\usepackage{url}
\usepackage{graphicx}
\usepackage{xspace}
\usepackage{subcaption}
\usepackage{booktabs}
\usepackage{wrapfig}
\usepackage{multirow}

\title{
\newtext{
\quest : Benchmarking Intrinsic Dimension Estimators on Homogeneous Manifolds with Anisotropic Embeddings
}}

\author{\name Aritra Das \thanks{Corresponding author.} \email aritrad@umd.edu \\
      \addr Joint Center for Quantum Information and Computer Science, \\ University of Maryland, College Park \& NIST. \\
      \AND
      \name Joseph T. Iosue \email jtiosue@gmail.com \\
      \addr Joint Center for Quantum Information and Computer Science, \\ University of Maryland, College Park \& NIST. \\ 
      Joint Quantum Institute, University of Maryland, College Park.
      \AND
      \name Victor V. Albert \email vva@umd.edu\\
      \addr Joint Center for Quantum Information and Computer Science, \\ University of Maryland, College Park \& NIST.
      }

\newcommand{\quest}{HomID\xspace}
\newcommand{\questfull}{\textbf{Hom}ogeneous Spaces for \textbf{I}ntrinsic \textbf{D}imension Estimation\xspace}
\def\month{12}  %
\def\year{2026} %
\def\openreview{\url{https://openreview.net/forum?id=aBJAPDsNvm}} %
\newcommand{\newtext}[1]{#1}

\newcommand{\U}{\mathrm{U}}
\renewcommand{\O}{\mathrm{O}}
\newcommand{\SO}{\mathrm{SO}}

\newcommand{\St}{\mathrm{St}}
\newcommand{\Gr}{\mathrm{Gr}}

\newcommand{\parentheses}[1]{\left(#1\right)}

\newcommand{\curlybrackets}[1]{\left\{#1\right\}}
\newcommand{\set}{\curlybrackets}

\newcommand{\pargs}[1]{\!\parentheses{#1}}

\newcommand{\braket}[2]{\left\langle #1, #2 \right\rangle}

\newcommand{\abs}[1]{\left\lvert #1 \right\rvert}

\newcommand{\bbC}{\mathbb{C}}

\newcommand{\bbI}{\mathbb{I}}

\newcommand{\bbR}{\mathbb{R}}

\newcommand{\bbZ}{\mathbb{Z}}

\newcommand{\calG}{\mathcal{G}}
\newcommand{\calH}{\mathcal{H}}

\newcommand{\calK}{\mathcal{K}}

\newcommand{\calM}{\mathcal{M}}
\newcommand{\calN}{\mathcal{N}}
\newcommand{\calO}{\mathcal{O}}

\newcommand{\calT}{\mathcal{T}}

\begin{document}

\maketitle

\begin{abstract}

The manifold hypothesis suggests that data lie on manifolds with smaller intrinsic dimension (ID) than their ambient dimension. However, there is no empirical agreement among ID estimates from different estimators for realistic datasets. Thus, it is important to test ID estimators (IDEs) with targeted stressors. In this work, we consider the role of anisotropy. To this end, we propose \quest, a collection of homogeneous spaces with anisotropic embeddings, for benchmarking ID estimators. We observe that methods that perform well on standard benchmarks systematically degrade on \quest under identical resource allocation. We further observe that anisotropic distortion of such benchmarks also results in performance degradation. Finally, we demonstrate that controlled anisotropic distortions induce systematic shifts in the distributions on which these methods rely, providing a concrete mechanism for the resulting estimation errors in two particular IDEs. \footnote{The code for this project is available at \url{https://github.com/a-writes-code/homid}
}

\end{abstract}

\section{Introduction}
\label{sec:intro}

The success of machine learning (ML) algorithms on datasets with large representative dimensions is often attributed to the hypothesis that the data actually lies on a manifold with smaller dimension, but embedded in a larger space \cite{Gorban2018_Manifold_Hypothesis, fefferman2013testingmanifoldhypothesis, Olah2014_MF, Cayton2005_MF}. This is called the manifold hypothesis.
An ML algorithm is able to generalize because it can infer this manifold (or sub-manifold) from the given training distribution. 
This notion is formalized by the concept of \textit{intrinsic dimension} (ID). 

The manifold hypothesis is quite intriguing and has been subject to many experimental investigations, whereby different methods have been proposed to estimate the ID of a particular data-cloud and applied to popular ML datasets such as MNIST \cite{lecun1998gradient,deng2012mnist}. 
However, a common theme is the disagreement in the estimated IDs of different methods on these real-world datasets
\cite{Hein2005Benchmark,Facco_2017_TwoNN,pope2021the_MNIST_10,20_bahadur2019dimensionestimationusingautoencoders,HiddenManifold, Candelori2025, MNIST_100_stanczuk2023diffusionmodelsecretlyknows, tempczyk2021lidl, kamkari2024a}.
This large variability between different methods suggests that either (i) the manifold hypothesis is incorrect, or (ii) these different methods have their inherent biases which may or may not be relevant to particular datasets. 

Existing benchmarks primarily rely on isotropic or weakly anisotropic structures, which we show can systematically overestimate estimator robustness. In this work, we undertake a careful study of how local anisotropy in data can cause systematic biases which degrade performance of several intrinsic-dimension estimators (IDEs). However, existing methods to induce anisotropy necessarily induce inhomogeneities in the data. To this end, we propose \textbf{\quest} (\questfull), a set of manifolds with homogeneous spaces and locally anisotropic embeddings known from classic differential geometry. Our proposed benchmark \newtext{highlights} the effect of local anisotropy, since the manifolds are otherwise homogeneous, have constant scalar curvature and have scalable embeddings. We summarize our core contributions below:
\begin{figure}
    \centering
    \includegraphics[width=0.8\linewidth]{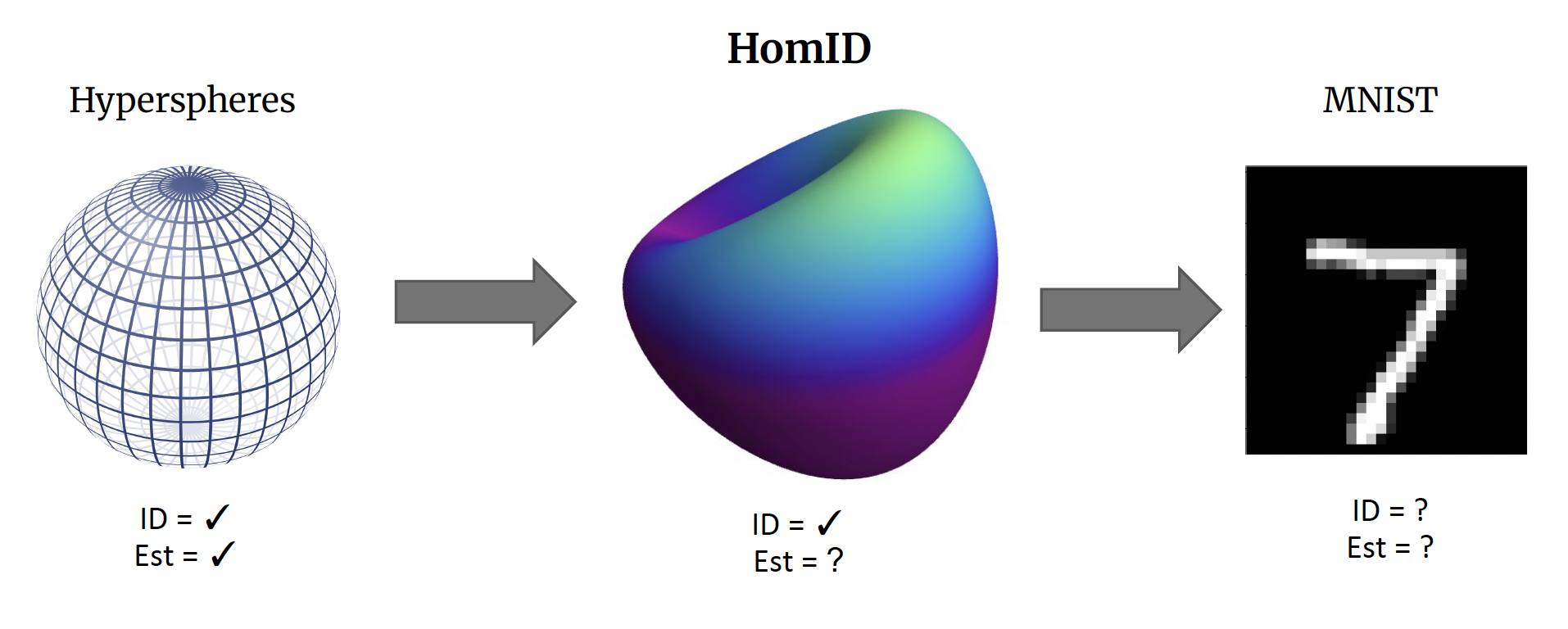}
    \caption{While most methods perform well when it comes to intrinsic dimension estimation (IDE) for simplistic manifolds like (hyper-)spheres, there's wide variability in their estimates for real-world datasets like MNIST.
    We propose benchmarking ID estimators with \quest---a family of manifolds known from classic differential geometry. \quest consists of homogeneous spaces with anisotropic embeddings, offering a clean way to understand the role of anisotropy in IDE evaluation.
    }
    \label{fig:intro}
\end{figure}

\begin{enumerate}
    \item We evaluate six ID estimators on \quest, a benchmark consisting of homogeneous spaces with anisotropic embeddings, derived using classic differential geometry results. We observe that all of these methods perform worse on \quest as compared to previous benchmarks except for a vector embedding of the Grassmannian manifold. \quest provides homogeneous, ground-truth manifolds whose embeddings exhibit strong local covariance anisotropy, and this stressor exposes failure modes not seen on comparable isotropic benchmarks. 
    \item We additionally stress-test IDEs by controlled anisotropic distortions of otherwise isotropic manifolds such as spheres and linear nullspaces, which also cause performance degradation. Thus, we show that it is important to move beyond simple isotropic datasets for benchmarking IDE methods.
    \item We analyze two particular ID estimators, namely lPCA and ABID, to test our hypothesis that anisotropy induces systematic biases in the underlying distributions, which lead to failure modes for the methods.
\end{enumerate}

\section{Background and Related Works}

\paragraph{Manifolds in ML} 
The manifold hypothesis states that real-world data lie on low-dimensional manifolds, even though expressed in high dimensions. \citep{Gorban2018Manifold, Cayton2005_MF}. This idea has been explored both theoretically and experimentally \citep{fefferman2013testingmanifoldhypothesis, HiddenManifold}. Topological properties of such data manifolds \citep{Carlsson2008} and latent representations \citep{ansuini2019intrinsicdimensiondatarepresentations} have also been studied.

\paragraph{Intrinsic Dimension} 
Intrinsic dimension (ID), as the name suggests, characterizes several inherent properties of the manifold \citep{Narayanan2009, pope2021the_MNIST_10}. A plethora of estimators have been proposed for ID estimation (IDE), which we discuss in Appx.~\ref{app:bg}. See \citep{CampadelliBench} for a detailed survey.

\paragraph{Benchmarks for ID Estimation} 

Practical benchmarks for IDE involve isotropic manifolds like hyperspheres, Gaussian distribution or non-scalable manifolds like the M\"obius strip 
\citep{CAMASTRA20032945, Hein2005Benchmark, Rozza2012, CampadelliBench, Bac_2021Scikit}.
An interesting benchmark proposed GAN-based data to lower bound ID \citep{pope2021the_MNIST_10}.
There also exist physics-inspired datasets \citep{santafe_weigend1993time, camastra2009comparative, GRASSBERGER1983189_Corrint} which typically use non-linear dynamical models. These benchmarks, however, do not isolate the effect of different stressors in a dataset for evaluating ID estimation. In particular, previous benchmarks \citep{tempczyk2021lidl, kamkari2024a} with anisotropic curvature typically have non-uniform density as well. %
\newtext{
\citet{tempczyk2026whyweneed} present a complementary argument for better LID benchmarking. They argue that existing evaluations can conflate domain- or architecture-specific inductive biases with true estimator performance. They therefore propose cross-domain benchmarks and controlled-transformation stress tests. 
} 
In this regard, we propose using homogeneous spaces as benchmarks to \newtext{investigate} the effect of anisotropy \newtext{in a controlled manner}.

\paragraph{Manifold constructions} 

Our methods are based on standard differential geometry, but
the specific recipes we use originate from the quantum optics and quantum information literature.
Namely, points in our manifolds correspond to Gilmore-Perelomov coherent states \citep{perelomov1977generalized-coh,perelomov1986generalized-coh,zhang1990coherent-states} --- the most ``classical'' states of a quantum system.
Such states are typically parameterized by the sphere or complex plane, but coherent states on more general manifolds are relevant to quantum information, quantum metrology \citep{holevo2011probabilistic} and, in the case of Grassmannians, the quantum Hall effect \citep{calixto2014coherent}.
As such, we sometimes refer to our construction as the ``coherent-state method''.
We also describe these quantum-mechanical recipes in the language of the ML community in Appx.~\ref{app:proof}.

A more comprehensive discussion of previous work may be found in Appx.~~\ref{app:bg}.

\section{Preliminaries}
\label{sec:prelim}

\paragraph{Manifolds and Intrinsic Dimension}
\label{ssec:manifolds}

A topological manifold $M$ of dimension $d_i$ is a topological space that locally ``looks'' like $\bbR^{d_i}$, in the sense that there exist local patches that are homeomorphic to $\bbR^{d_i}$ \citep{lee2012introduction-to}.
For a given disjoint union of manifolds and a point $p$ in this union, the local intrinsic dimension around $p$ is the dimension of the submanifold to which it belongs \citep{kamkari2024a}.

In addition to this \textit{intrinsic} viewpoint of a manifold, one can equivalently consider manifolds \textit{extrinsically} by defining them as appropriate subsets of some ambient Euclidean space $\bbR^{d_a}$ \citep{lee2012introduction-to}. 
Given (samples from) a set $S \subset \bbR^{d_a}$ along with the promise that $S$ is a manifold of dimension $d_i \leq d_a$ for some unknown ID $d_i$, one may naturally desire an algorithmic procedure to estimate $d_i$.
Intuitively, the ID of a dataset can be thought of as the minimum number of parameters needed to represent the data with no loss of information \citep{DANCOCERUTI20142569}.
Importantly, though, a $d_i$-dimensional manifold can have nontrivial geometry (such as a direction-dependent, or anisotropic, curvature) and topology (such as being not simply connected) that makes it starkly different from real space or a sphere of the same dimension.

Our manifolds, by definition, exhibit a single ID at all points.
In contrast, most methods return the local (scale-dependent) ID (LID) estimates at different points. Thus, we will refer to the \textit{mean} of these LID estimates as \textit{the} ID of the manifold as estimated by a given method. \footnote{We did not observe any significant differences with other statistical quantities such as using median, mode, mean of medians and median of means. This indicates that methods are not returning too much of variation for the IDE values.}

\paragraph{Homogeneous spaces}
\label{ssec:qs}

All manifolds included in the \quest benchmark are parameterized by quotient spaces \(\cal G/H\) (a.k.a. homogeneous spaces), where \(\cal G\) is a Lie group (i.e., a group that can also be considered as a manifold), and where \(\cal H\) is one of its subgroups. 
For \(\cal G\), we use either the orthogonal group \(\O(n)\) or the unitary group \(\U(n)\), which is the symmetry group of the \(n\)-dimensional real or complex sphere, respectively.
We provide a pedagogical flavor of these spaces below, leaving technical details to Appx.~~\ref{app:proof}.

The two-dimensional real sphere, \(S^2\), is a simple example of a homogeneous space. 
There exists a proper (i.e., orientation-preserving) rotation that can map the north pole to any other point on the sphere (whose action can be thought of as moving along the great circle connecting the north pole to the desired point).
Therefore, we can label all points on the sphere by the rotations that take us there from the north pole, but we need to omit all rotations that rotate around the north pole since those do not take us anywhere.

Mathematically, this translates to the homogeneous space \(\text{SO}(3)/\text{SO}(2) \cong S^2\), where \(\text{SO}(3)\) is the group of all proper three-dimensional rotations, and \(\text{SO}(2)\) is the subgroup of rotations around the north pole.

The characteristics of the quotient space depend on the subgroup, which can be continuous or discrete.
Spheres and more general homogeneous spaces whose \(\cal H\) is a continuous group are constructed in the same spirit as this example.
Their intrinsic dimension is \(d_i = \dim{\cal G} - \dim \cal H\). 

\paragraph{Local Anisotropy}
\label{ssec:anisotropy}

\newtext{
There exist several notions of anisotropy in the literature. We enlist some of them below.
\begin{itemize}
    \item Representation learning: A point could is called anisotropic if its \textit{global} covariance concentrates in a few directions. \cite{ethayarajh2019contextual, rudman2022isoscore}. Appendix~\ref{app:null} reports null correlation between global covariance and IDE performance, which necessitates thinkinga about a local measure of anisotropy. We note that this distinction between local and global covariances is not unique and was also explored, for instance in \cite{cai2021isotropy}.
    \item Intrinsic notion in differential geometry: Anisotropy in this context refers to symmetry along different directions. \quest manifolds are anisotropic in this sense, since compact spaces that are both homogeneous and isotropic are completely classified. These are spheres and some projective spaces listed in \cite{wang1952two}.
    \item Extrinsic notion in differential geometry: A surface sitting inside $\mathbb{R}^{d_a}$ can also have anisotropic bending along different tangent directions. This is captured by the second fundamental form $\mathbb{I}(u,v)$. A surface is called locally umbillic (navel-like) at a point $p$ if the magnitude of $\mathbb{I}(u,u)$ at a point $p$ is the same for every direction $u$. Spheres and flat planes are the only surfaces which are umbillic everywhere in Euclidean plane, a result which captures the canonical notion of anisotropy in this regard \cite{oneill1965isotropic}.
\end{itemize}
}

\newtext{
In our work, we propose the local covariance spectra as a convenient proxy for measuring the local anisotropy. This measure can be confounded by the effects of curvature and sampling density. However, \textit{\quest, by construction, has the same density and geometry at every point under invariant sampling.} In addition, by a symmetry argument (which we outline in Appendix~\ref{app:anisotropy}, one can show that for \quest manifolds, in the limit $k,N \to \infty, k/N \to 0$, the local covariance spectra consists of \textit{flat plateaus arranged in a staircase-like fashion}.
Deviations from these plateaus can arise from finite-$k$ and finite-$N$ effects. 
This follows from the fact that for any smooth manifold \cite{singer2012vector,kaslovsky2014nonasymptotic,little2017multiscale,tyagi2013tangent,alvarezvizoso2020local,pottmann2007principal}, the local covariance spectra has the form 
\begin{equation}
    \label{eq:lpca-form}
    C(\epsilon) \approx \frac{\epsilon^2}{d_i + 2} \Pi_\textbf{tangent} + \epsilon^4 Q(\mathbb{I}) + o(\epsilon^4)
\end{equation}
where $\Pi_\textbf{tangent}$ is the projector onto the tangent space and $Q(\mathbb{I})$ is a quadratic form involving the second fundamental form. The former represents the (dominant) flat or isotropic contribution, and the latter represents the anisotropic contribution. Thus the local covariance spectra is necessarily a scale-dependent feature.
In addition, this implies that the signal for anisotropy is present in the tail of the spectrum.
}

\newtext{
\quest embeddings are known to be anisotropic from classic results in differential geometry \cite{wang1952two, helgason1978differential, wolf2011locally, besse2008homogeneous}. In addition, we provide independent numerical verification in Appendix~\ref{app:anisotropy}, where we compare \quest manifolds to matched spheres with the same intrinsic and ambient dimensions. We observe approximately flat bands as expected, with the isotropic sphere exhibiting a strong degradation beyond its intrinsic dimension. In contrast, \quest manifolds  carry non-trivial signatures of their anisotropy in the spectral tail.
We present this as numerical evidence for the anisotropy of \quest manifolds, but do not claim this to do an exact characterization of estimator failure modes.
}

\section{Proposed Datasets}

\newtext{
The five \quest configurations probe related but distinct aspects of anisotropic embedding geometry. The two Stiefel embeddings, 'St (Matrix)' and 'St (Vec)', are the closest extensions of spheres. They implicitly test whether anisotropy degrades estimator performance even for this mild homogeneous-space generalization.
The two Grassmannian configurations, 'Gr (Proj)' and Gr '(Vec)', use closely related quotient spaces with the same intrinsic dimension, but different embedding maps. In particular, 'Gr (Proj)' uses projector vectorization, whereas 'Gr (Vec)' uses the Pl\"ucker-type coherent-state embedding.
The ``Flag (Vec)'' embedding extends the Grassmannian construction to a coarser partition of subspaces. It tests whether anisotropy effects become stronger in a higher-order composite quotient.
The current family of manifolds can be easily extended to other homogeneous-space families. Examples include quotients by finite subgroups and quotients involving symplectic or complex unitary groups. Section~\ref{sec:conclusion} discusses some of these extensions.
}

\paragraph{Stiefel manifolds}
\label{ssec:st}

Stiefel manifolds are the closest relatives of spheres out of all \quest manifolds.
Real Stiefel manifolds are parameterized by quotients of the form $\O(n)/\O(n-k)$ for \(n\geq k \geq 1\) \citep{james1977the-topology-of}, while real spheres can also be understood as the quotients \(\text{O}(n)/\text{O}(n-1)\). 

A common alternative definition is the set of \(n\times k\) real matrices \(X\) satisfying \(X^T X = \bbI\), where \(T\) is the transpose map, and where \(\bbI\) is the appropriately sized identity matrix.
In this way, one can interpret the manifolds as all possible isometries of \(k\)-dimensional space into \(n\) dimensions.
We note that the topology of general Stiefel manifolds is quite different from that of spheres \citep{james1977the-topology-of}. 

The \quest benchmark includes two different embeddings of Stiefel manifolds.
The first, called ``St (Matrix)'', is a simple vectorization of the matrices \(X\).
The second, called ``St (Vec)'', is a mixture of the coherent-state method and a vectorization of a matrix.

\paragraph{Grassmannians}
\label{ssec:gr}

Grassmannians, or Grassmann manifolds, are defined as $\Gr(k,\bbR^n) \cong \O(n)/\O(n-k) \times \O(k)$ \citep{lee2012introduction-to} and can be thought of as quotients of Stiefel manifolds by an extra \(\O(k)\) subgroup. 
Grassmannians have a long history in ML \citep{zhang2018grassmannianlearningembeddinggeometry,bendokat2020a-grassmann-man}.
A simple example of one is the real projective plane, \(\bbR P^{2} \cong \Gr(1,\bbR^3)\), which is the sphere \(S^2\) with antipodal points identified.

The \quest benchmark includes two different embeddings of Grassmann manifolds.
The first is based on the interpretation of the Grassmannian as a space of subspaces.
The quotient of the Stiefel manifold by the extra \(O(k)\) subgroup identifies two isometries related by a basis change as equivalent.
This implies that Grassmannians parameterize all distinct \(k\)-dimensional subspaces of \(n\)-dimensional space.
We represent a point on the Grassmannian by an \(n\)-dimensional projector onto a $k$-dimensional subspace, which yields our ``Gr (Proj)'' embedding when written as an \(n^2\)-dimensional vector.

The second embedding, called ``Gr (Vec)'', is based on the coherent-state method and allows for an ambient dimension as low as \(\binom{n}{k}\).
It is also known as the Pl{\"u}cker embedding \citep{lee2012introduction-to,bendokat2020a-grassmann-man}.

\paragraph{Flag manifolds}
\label{ssec:flag}
Flag manifolds generalize Grassmannians to arbitrary \textit{sets} of $n$-dimensional vectors. 
A $t$-flag manifold is defined as the manifold described by the quotient space $\O(n) / \O(k_1) \times\O(k_2)\times \cdots\times \O(k_{t+1})$, with the constraint $\sum_{i=1}^{t+1} k_i = n$. 
Our benchmark contains the ``Flag (Vec)'' embedding of the \(t=2\) case, which is based on the coherent-state method.

\section{Methodology}
\label{sec:methodology}

We briefly summarize the sampling procedure and evaluation protocol here; full details are in Appx.~\ref{app:proof} (manifold constructions) and Appx.~\ref{app:exp-hprm} (experimental details). We will publicly release the code upon acceptance.

\newtext{
\paragraph{Sampling procedure} Each of the five \quest embeddings starts from a Haar-random element of the ambient orthogonal group $\O(n)$. We then apply the corresponding deterministic embedding map from Appx.~\ref{app:proof} --- for 'Gr (Proj)', we use a projector; for 'St (Matrix)', we use the leading $k$ columns; for 'Gr (Vec)', we use the vector of $k \times k$ minor determinants, which gives a Pl\"ucker-type embedding. For 'St (Vec)', we concatenate this vector with an independently sampled $V \in \O(k)$. For 'Flag (Vec)', we use the outer product of two minor-determinant vectors. Each of the $N$ points in a manifold configuration uses an independent random draw.
}

\paragraph{Methods tested}
We choose a few standard representative IDE methods of different flavors for testing on our benchmark.
These are linear methods like linear subspace projection (lPCA \citep{PCA_Cangelosi,PCA_FO, fan2010intrinsicdimensionestimationdata_PCA}) as well as non-linear methods based on maximum likelihood estimation (MLE \citep{Levina_MLE, Haro2008MLE}), fractal dimension estimation (CorrInt \citep{GRASSBERGER1983189_Corrint, 1039212_CorrInt}), distribution of distances (TwoNN \citep{Facco_2017_TwoNN}), angles (ABID \citep{THORDSEN2022101989}) as well as concentration of measure (DANCo \citep{DANCOCERUTI20142569}). %
We review these methods, including their implementation, in Appx.~\ref{app:std-skdim}.
\newtext{Unless otherwise noted, lPCA results use the `maxgap` variant.}
\newtext{
We emphasize that this list is not exhaustive. For our experiments, we use the following three selection criteria. First, the methods represent a diverse set og methodological families. The selected methods include subspace/PCA-based (lPCA), likelihood-based (MLE), fractal/correlation-integral-based (CorrInt), nearest-neighbor-distance-based (TwoNN), angle-based (ABID), and concentration-of-measure-based (DANCo) approaches. Second, the methods are widely used for IDE. This criterion makes the results relevant to methods that practitioners commonly use. Third, each method has a well-tested open-source implementation. We also choose methods such that the runtimes are tractable for the sample sizes used in our sweeps, up to $N=10^4$.
}

\newtext{
\paragraph{Default hyperparameters} The most substantial difference in ID estimates comes from the hyper-parameters $k$ and $N$, which we checked in several short-scale experiments. Thus, we use method-specific defaults $\sigma=0.1$ for MLE, a discard fraction of $0.1$ for TwoNN, and $(k_1,k_2)=(10,20)$ for CorrInt for the experiments presented in this paper. Details are provided in Appx.~\ref{app:exp-hprm}.
}

\newtext{
\paragraph{Seeds and aggregation} We use two types of seeds. A manifold-generation seed is fixed once for each manifold configuration. An experiment seed controls the random subsample of $N$ points in each run and any additive noise. For each experiment seed, we compute a pointwise local ID estimate at every sampled point. We then average these estimates to obtain the run-level estimate $\hat{d}_i$ (Sec.~\ref{ssec:manifolds}). All reported values are averages over three experiment seeds.
Two sources of variation are present in our experiments. The first is variation across local ID estimates at different sample points. The second is variation across the three experiment seeds. The displayed standard deviations measure pointwise heterogeneity in the ID estimate in a given sample. We compute the within-seed standard deviation of the local estimates for each run and then average these standard deviations across the three runs.
}

\section{Results}

\subsection{IDE Performance on \quest}

Given a manifold embedded into a space with ambient dimension \(d_a\), we define the \textit{relative error} $\delta$,
\begin{equation}
    \label{eq:rel-err-def}
    \boxed{\delta:=\frac{\hat{d}_{i}}{d_{i}}-1\in \left[-1, \frac{d_a}{d_i} - 1 \right]}\qquad\text{(relative error)}
\end{equation}
Here, $\hat{d}_i$ is an estimated quantity, while $d_i$ is the ground-truth ID. 
Note that $\delta < 0$ implies that the method underestimates the ID, while $\delta > 0$ indicates an overestimation for the ID up to the ambient dimension. 
Average performance of methods on \quest manifolds is summarised in Table~\ref{tab:perf}, in which we list \(|\delta|\) so as to compare over- and under-estimation on the same footing.  \newtext{Unless otherwise noted, a worse performance corresponds to a higher absolute value of $\delta$.}

\begin{table*}[t!]
\centering
\resizebox{0.99\textwidth}{!}{
\begin{tabular}{||l|ccccc||c||}
\toprule
\textbf{IDE Method} & \shortstack{Gr\\(Proj)} & \shortstack{Gr\\(Vec)}  & \shortstack{Flag\\(Vec)} & \shortstack{St\\(Matrix)} & \shortstack{St\\(Vec)} & \bf{\shortstack{Average\\(\quest)}} \\

\midrule
lPCA & $0.1725 \pm 0.1544$ & $0.0926 \pm 0.0195$ & $0.2068 \pm 0.0586$ & $0.0666 \pm 0.1153$ & $0.0666 \pm 0.1154$ & $0.1210 \pm 0.1192$ \\
MLE & $0.1399 \pm 0.0280$ & $0.1716 \pm 0.0255$ & $0.2555 \pm 0.0452$ & $0.2003 \pm 0.0591$ & $0.1852 \pm 0.0540$ & $0.1905 \pm 0.0586$ \\
CorrInt & $0.7025 \pm 0.0100$ & $0.7141 \pm 0.0085$ & $0.8221 \pm 0.0051$ & $0.7661 \pm 0.0077$ & $0.7632 \pm 0.0079$ & $0.7536 \pm 0.0434$ \\
TwoNN & $0.0964 \pm 0.0070$ & $0.0913 \pm 0.0120$ & $0.1643 \pm 0.0382$ & $0.1110 \pm 0.0420$ & $0.1121 \pm 0.0375$ & $0.1150 \pm 0.0405$ \\
ABID & $0.1258 \pm 0.0362$ & $0.1316 \pm 0.0333$ & $0.3712 \pm 0.0475$ & $0.2099 \pm 0.0472$ & $0.1995 \pm 0.0432$ & $0.2076 \pm 0.0981$ \\
DANCo & $0.1923 \pm 0.0187$ & $0.1371 \pm 0.0235$ & $5.4662 \pm 0.0402$ & $0.1337 \pm 0.0290$ & $0.1462 \pm 0.0188$ & $1.2151 \pm 2.1258$ \\
\midrule
\textbf{Average} & $0.2382 \pm 0.2202$ & $0.2230 \pm 0.2224$ & $1.2144 \pm 1.9145$ & $0.2479 \pm 0.2445$ & $0.2455 \pm 0.2427$ & $0.4338 \pm 0.9637$ \\
\bottomrule
\end{tabular}%
}
\vspace{0.08in}
\caption{Mean relative error $|\delta|$ for various methods on \quest manifolds. A higher value indicates worse performance. We see that the vector embedding of Grassmannian consistently has low error for all methods, while TwoNN typically performs the best on all manifolds. The quantity after $\pm$ is the pooled within-seed standard deviation of the pointwise local relative errors; it measures spatial dispersion of local estimates, not uncertainty in the reported mean.}
\label{tab:perf}
\end{table*}

\subsection{Comparison with other Benchmarks}
\label{ssec:iso-homid-comp}

\begin{figure}[t!]
    \centering
    \includegraphics[width=0.95\linewidth]{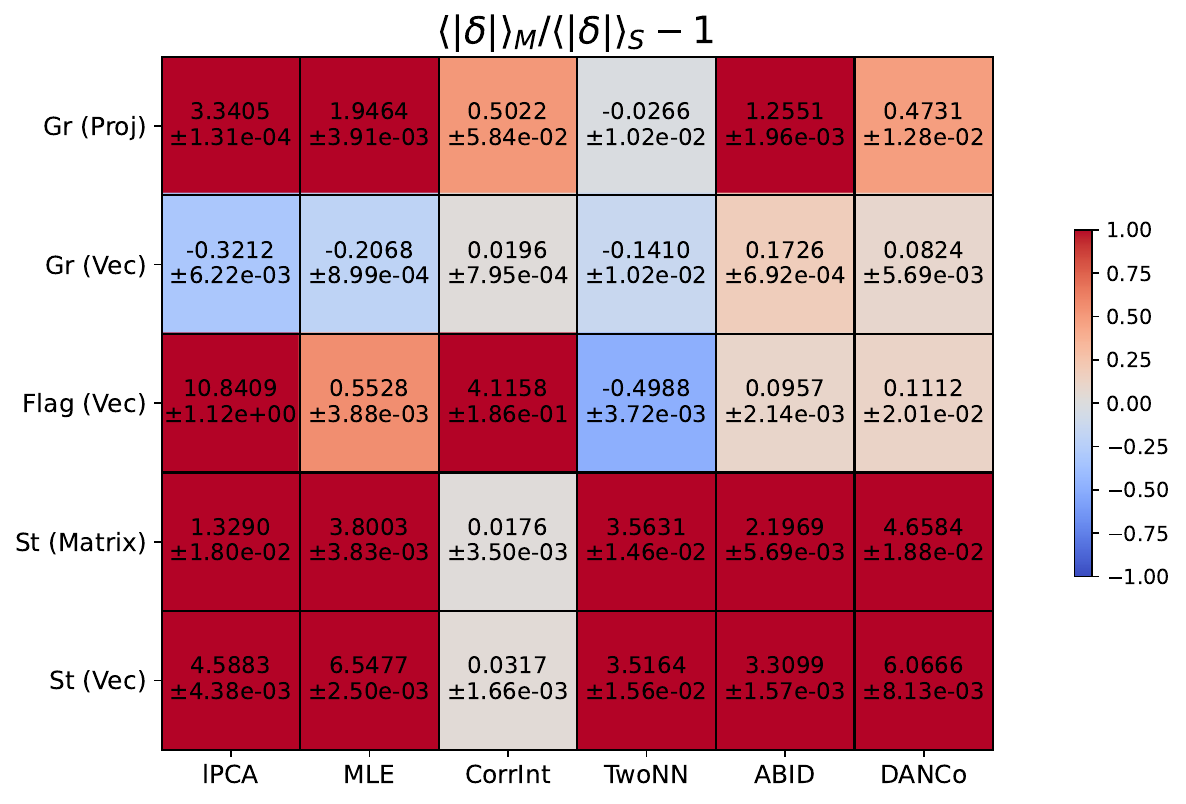}
    \caption{
    Comparison of the quantity \(\langle |\delta|\rangle_M / \langle |\delta|\rangle_S - 1\), where the numerator is the average of the absolute value of the relative error \(|\delta|\) over all instantiations of a given manifold family \(M\), while the denominator is the corresponding average over all sphere embeddings with the same intrinsic and ambient dimensions.
    This relative comparison shows that tested methods tend to perform much \textit{worse} against our manifolds than against spheres with the same dimensions. 
    Interestingly, we observe a positive score with a change in embedding of the Grassmannian, \newtext{which we defer to understanding in more detail for future work.}
    Due to high computational time, we choose a smaller range of hyper-parameter sweeps for DANCo. A 1-$\sigma$ sampling error is reported here after the \(\pm\) sign. 
    \newtext{Note that the color scale is clipped at 1.0 for ease of visualization. Larger values indicate severe degradation relative to spheres.}
    }
    \label{fig:perf-hs}
\end{figure}

We undertake a comprehensive evaluation of IDE methods on \quest. We perform hyper-parameter sweeps and, for each hyper-parameter combination, compare the performance of different IDE methods on \quest to other IDE benchmarks with isotropic manifolds. 
We notice that the chosen methods are almost always worse at estimating the ID for our manifold embeddings, with the notable exception being the embedding ``Gr (Vec)''. For brevity, we present the relative result on spheres here in the main text, cf. Fig \ref{fig:perf-hs}. \newtext{Matched spheres are our primary baseline since they have scalable and arbitrary intrinsic and ambient dimensions, allowing us to compare them with \quest manifolds with identical resource allocation. In addition, it helps conflate the scale-dependency problem we discussed in Section~\ref{ssec:anisotropy}.}
\newtext{
We observe similar qualitative results for other isotropic manifolds, including isotropic Gaussian spaces and linear affine spaces. Appx.~\ref{app:bm} gives the detailed plots.
}
\newtext{
We restrict the comparison to manifolds with scalable and freely tunable intrinsic and ambient dimensions under matched sampling resources. This requirement excludes benchmarks such as the Möbius strip, torus, and Swiss roll.
}
We emphasize that the comparison is made across a range of scales, accessible by our computational resources.%

\subsection{Stress-testing the role of anisotropy}
\label{ssec:aniso}

The hypothesis that anisotropy limits robustness of IDE methods suggests that anisotropic distortion of isotropic manifolds should also cause degradation of IDE performance. We perform two sets of controlled experiments to obtain such distortions.

\paragraph{Random diagonal transformations}
\label{para:squeeze}

We test IDE methods on anisotropic distorted versions of \quest manifolds, naturally obtained by amending the coherent-state method by linearly transforming the data with a random diagonal matrix. Such distorted manifolds have locally anisotropic curvature, but in addition they get non-uniform curvature and density. However, the ID is the same.

\begin{figure}[t]
    \centering
    \includegraphics[width=0.95\linewidth]{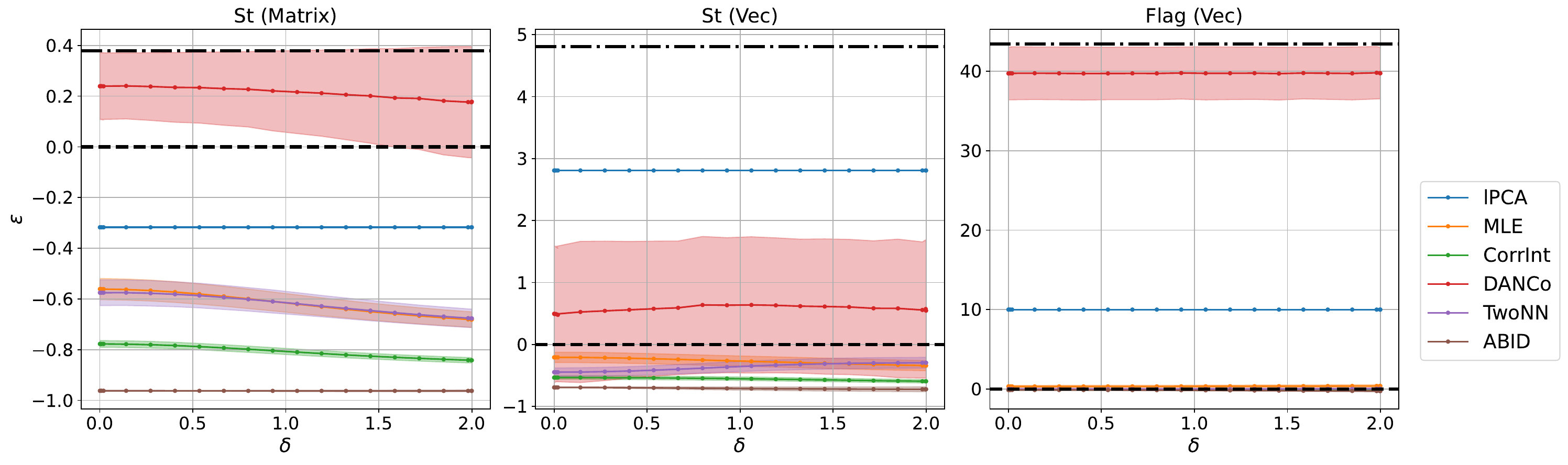}
    \caption{Effect of anisotropic distortion: We plot the relative error $\langle\delta\rangle$ as a function of the parameter $\epsilon$, which is a direct measure of anisotropy. Except for St (Matrix), most methods show negligible change as $\epsilon$ is increased.}
    \label{fig:squeeze}
\end{figure}

The strength of distortion is governed by a parameter $\epsilon$, and we generate each diagonal entry by sampling uniformly from $[1- \frac{\epsilon}{2},1 + \frac{\epsilon}{2}]$. The performance of methods is mostly unchanged upon distortion of the underlying manifolds. 
We demonstrate this for three typical manifolds in Fig~\ref{fig:squeeze}, adding the rest to Appx~\ref{app:add-squeeze-noise}.
Some methods show a slight degradation in performance, but the change is minimal and within the error margin. By contrast, the methods performed significantly worse on distorted spheres than on our distorted manifolds, cf. App~\ref{app:squeeze-compare-spheres}.

\paragraph{Additive noise}
\label{para:noise}

An alternate way to understand the role of anisotropy is to perturb the original manifolds with additive noise, i.e. $\mathbf{x} \to \mathbf{x} + \bf{\epsilon}$, where $\bf{\epsilon}$ is sampled from a gaussian distribution $\calN(\mathbf{0}, \mathbf{\Sigma})$. The role of anisotropy can be clearly understood by choosing $\mathbf{\Sigma}$ to be either the identity matrix $\mathbf{I}_{d_a}$ (\textit{isotropic}), a diagonal matrix $\Lambda$ with elements chosen from $\mathcal{U}(0,\frac{2}{d_a})$ (\textit{uncorrelated}), or a random positive definite matrix $u u^T$ (\textit{anisotropic}), where the elements of $u \in \mathbb{R}^{d_a \times d_a}$ are chosen from $\mathcal{N}(\mathbf{0}, 1)$. We then explicitly set $\text{Tr } \Sigma = 1$ by the transformation $\Sigma \to \Sigma' = \Sigma / (\text{Tr } \Sigma)$.

Most methods deviate in their estimations when $\sigma^2 \sim ||\mathbf{x}||_2^2 = \mO(1)$.
Rather curiously, we observe flat lines, indicating that, for certain methods, the data cloud is indistinguishable from pure noise. 
We also observe an improvement in IDE performance for certain manifold-method combinations, suggesting a certain regularizing effect emerging from the noise. 

We observe that in the low to intermediate regime, there is no discernible change in the behavior of isotropic or anisotropic noise. However, in the high noise limit, we observe that anisotropic noise is always underestimated, further corroborating our conjecture.

\begin{figure*}[b!]
    \centering
    \includegraphics[width=0.95\linewidth]{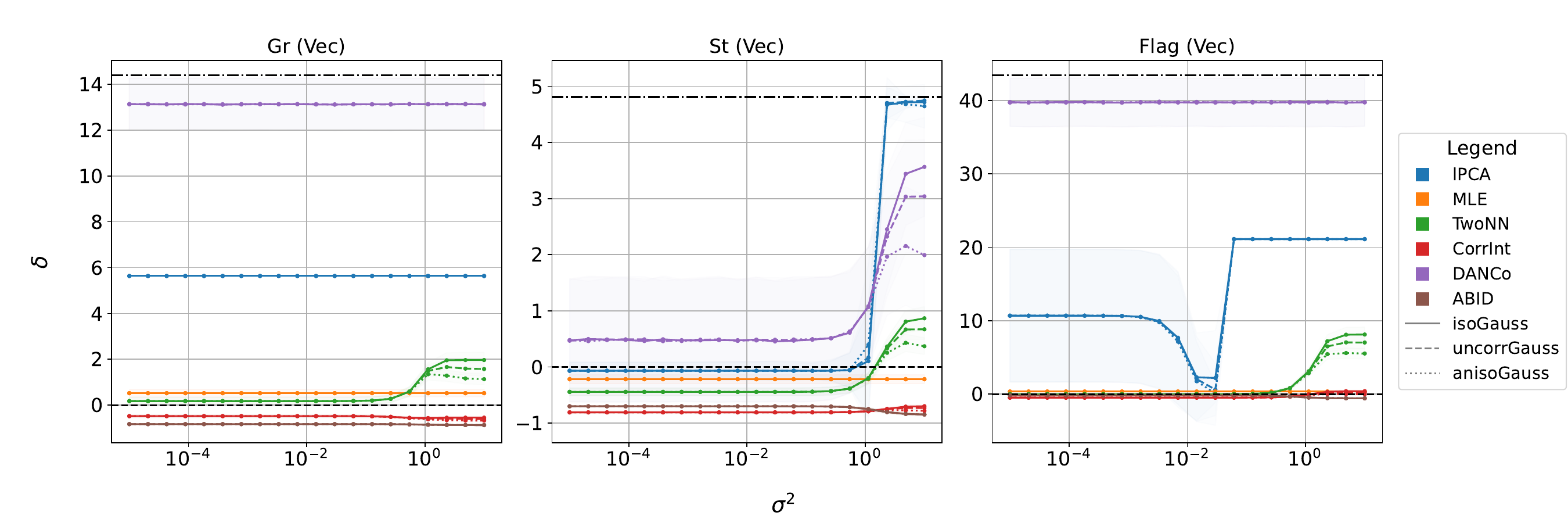}
    \caption{Effect of additive noise. We report IDE performance when the unperturbed data $\mathbf{x} \to \mathbf{x} + \epsilon$ where $\epsilon \sim \mathcal{N}(\mathbf{0}, \sigma^2 \Sigma)$ with $\Sigma$ chosen to be proportional to the identity (isotropic), diagonal (uncorrelated) or a  random symmetric matrix (anisotropic).  We then plot the relative error $\delta$ (on the true manifold) as a function of the noise scale $\sigma^2$.
    We observe all three types of behavior when noise is added, as discussed in Section~\ref{para:noise}.
    Notice that there is no discernible change in behavior between the different noise types, except in the high noise limit, where the anisotropic noises are consistently underestimated.%
    A 1-$\sigma$ sampling error is plotted.
    \\
    }
    \label{fig:anisoNoise}
\end{figure*}

The results are summarized in Fig.~\ref{fig:anisoNoise}. We present three typical manifolds here, with plots for all the manifolds in Appx.~\ref{app:add-squeeze-noise}.

\subsection{Relation to real-world data}
\label{ssec:mnist}

\begin{figure}[h]
    \centering
    \includegraphics[width=\linewidth]{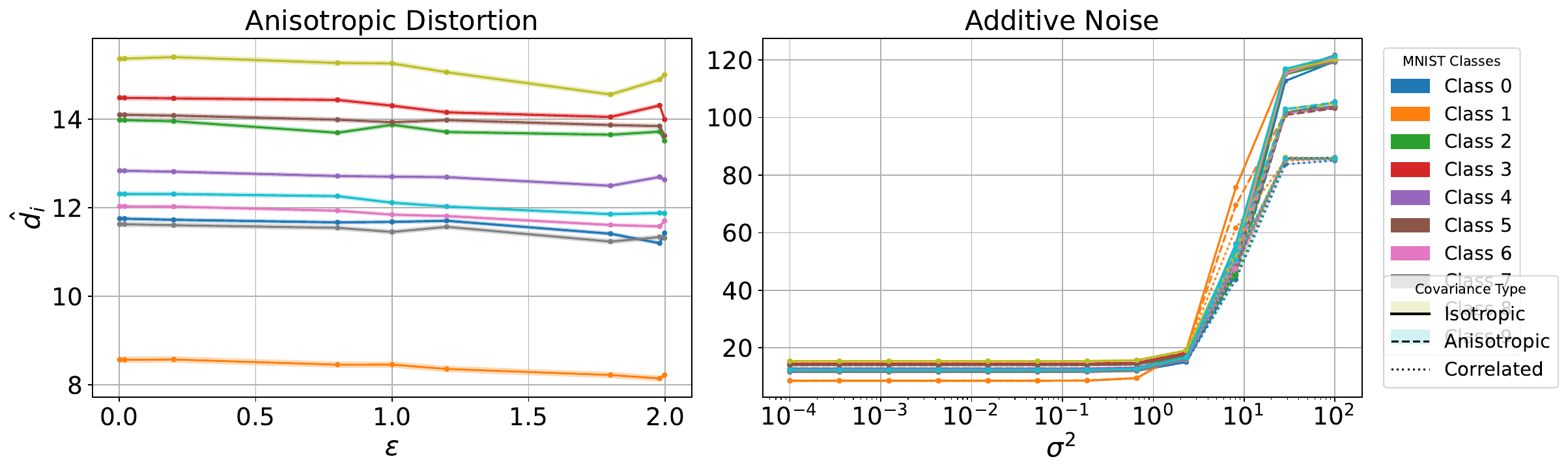}
    \caption{IDE performance for MLE on MNIST when subjected to anisotropic distortion and noise perturbation. Notice how additional anisotropy does not change the IDE by much, similar to \quest. The additive noise alters the prediction only at $\sigma^2 \sim \mO(1)$. \newtext{The other plots, where we observe qualitatively similar behavior, are presented in Appx.~\ref{app:mnist}.}}
    \label{fig:mnist-mle}
\end{figure}

As a sanity check, we performed the above anisotropic distortion experiments with \newtext{image datasets MNIST, FashionMNIST and CIFAR-10}. We show in Figure~\ref{fig:mnist-mle} how these distortions yield results that are qualitatively similar to that of \quest for a representative method-dataset pair. This highlights the importance of benchmarking IDE methods with such controlled stressors which can be readily applied to realistic datasets. Note that this experiment is not validation of exact quantitative transfer, but rather that the anisotropy stressor is not artificial.

\subsection{\textit{Why} does anisotropy hurt?}

To leading order, the Euclidean and geodesic distances between two nearby points differ because of the curvature \citep{lang1995differential-an}.
For manifolds with anisotropic curvature, this difference depends on the points. 
Nearest-neighbor (NN) IDE methods use Euclidean distances to choose the $k$NNs of a point on the manifold, which can cause systematic bias in the choice of neighborhoods (assuming finite neighbors $k$ and sample size $N$). 
\newtext{
These neighborhood distortions can introduce systematic bias into kNN-based estimators at finite k and N. The \quest and controlled-distortion experiments provide empirical evidence consistent with this mechanism.
}

\subsection{Analysis of other statistical and geometric properties}
\label{ssec:null}

We find that commonly used scalar quantities are insufficient to explain IDE performance differences. These include properties such as global variance, global anisotropy, correlations between different Euclidean components of the vectors, local density and local curvature. 
\newtext{
We found no statistically significant correlation between these properties and IDE performance. Appx.~\ref{app:null} provides the full details of these test, including the algorithms used to measure the quantities.
}
\newtext{
We emphasize here that we do \textbf{not} claim a complete causal identification of local anisotropy for explaining IDE failure. However, the null results presented here, along with the complementary controlled experiments presented in Section~\ref{ssec:iso-homid-comp} and Section~\ref{ssec:aniso} provide triangulating evidence that local anisotropy is correlated with IDE errors. We leave an in-depth characterization of the failure modes for future work.
}

\section{Failure mode analysis for lPCA and ABID}

We analyze the reasons for failure for two IDE methods in this section. Most methods usually rely on some assumptions about the underlying data distribution if they belong to a manifold. We show how anisotropy \newtext{is associated with} deviations from such distribution for the lPCA and ABID methods. Note that an exhaustive quantitative treatment of this effect is beyond the scope of this project.

\begin{figure}
    \centering
    \includegraphics[width=0.9\linewidth]{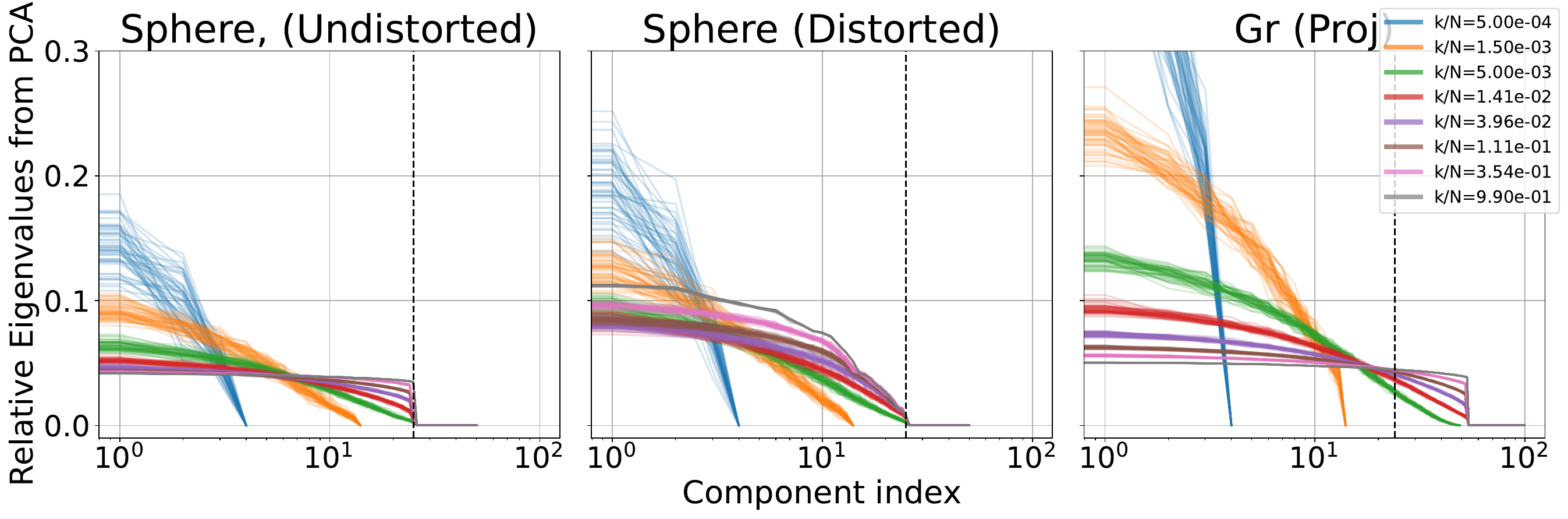}
    \caption{Analysis of relative eigenvalues for lPCA. Recall that the relative eigenvalues represent the fraction of total variance explained by top PCA components. We observe that the relative eigenvalues show a sharp decline at the correct ID value for undistorted spheres, the kink disappearing with distortion. However, for Grassmannian projectors, we observe that the sharp drop does happen but at a different ID. Dashed black lines represent the ID of the manifold.}
    \label{fig:lpca-interp-mt}
\end{figure}

We conjecture that lPCA fails in the presence of anisotropic noise due to deviation from the expected distribution for manifolds.
Recall that lPCA looks at the directions of maximal variance in a local neighborhood. For a manifold, the normal directions show zero variance. Thus we expect a sharp drop in the relative eigenvalues at the index corresponding to the ID.  In Fig.~\ref{fig:lpca-interp-mt}, we observe that spheres show this behavior, but distorted spheres and \quest manifolds do not. 
In particular, this gap sharply decreases as we distort the sphere, while the Grassmannian exhibits a drop at a different index. This shows that the failure mode for lPCA arises from a decreasing signal-to-noise ratio for distorted spheres, while for \quest, the failure mode is because of an incorrect signal. This trend, however, does not occur exactly for all \quest manifolds, which we describe in detail in Appx.~\ref{app:lpca}.

\begin{figure}[b!]
    \centering
    \includegraphics[width=0.9\linewidth]{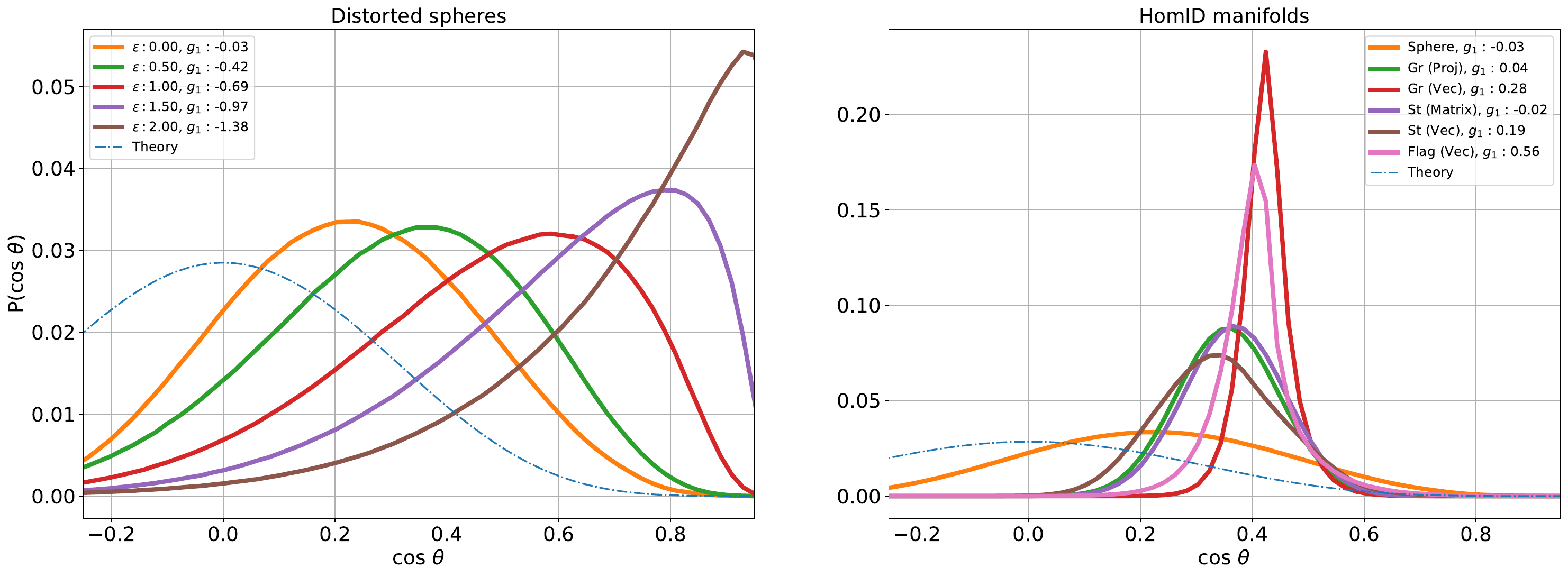}
    \caption{Cosine distribution for ABID: The distribution of cosine similarity between random unit vectors is expected to follow a Beta distribution centered at zero (dashed-dotted curve in blue). In practice due to correlations between the vectors chosen in a $k$-local neighborhood, there is a slight shift, but the overall symmetry is preserved. As we distort a sphere (left subpanel) by increasing $\epsilon$, we notice a gradual change in asymmetry, which affects the ID estimate. This is reflected by the increased value of the skewness $g_1$ for the different samples. A similar behavior is seen for the \quest manifolds (right subpanel). 
    }
    \label{fig:abid-interp}
\end{figure}

We also analyze the failure mode for ABID. As implemented in \citep{THORDSEN2022101989}, the distribution of pairwise cosine similarities between independent random vectors on a $d_i$-dimensional hypersphere is expected to follow a Beta distribution. ABID calculates cosine similarity between vectors in a local neighborhood. Thus the distribution is slightly shifted, due to correlations, but still overall symmetric. This variance controls the dimension estimate, as outlined in Appx.~\ref{app:std-skdim}. With increasing anisotropy, we observe a gradual asymmetry in the distribution, which leads to an incorrect estimate. This is true for both distorted spheres as well as \quest datasets, cf. Fig.~\ref{fig:abid-interp}. 

We believe that anisotropic curvature leads to similar systematic biases in distributions relevant for other methods. 
\newtext{
We summarize these distributional shifts in Appendix~\ref{app:mech-interp}.
}

\section{Conclusion, Limitations and Future Work}
\label{sec:conclusion}

\paragraph{Conclusion} In this work, we investigated the role of anisotropy as a stressor for intrinsic dimension estimation (IDE). We introduced a new benchmark family, \quest, consisting of homogeneous manifolds with anisotropic embeddings and known ground-truth intrinsic dimension.
Across a range of commonly used estimators, we found that methods that perform well on standard synthetic benchmarks—such as hyperspheres—consistently degrade on \quest under matched sampling conditions. Furthermore, controlled anisotropic distortions of classical benchmarks reproduce similar degradation, suggesting that isotropy in existing benchmarks masks important failure modes.
Our analysis of representative estimators (lPCA and ABID) indicates that anisotropy induces systematic biases in the distributions these methods rely on, leading to incorrect dimension estimates. While this does not fully characterize all failure mechanisms, it provides concrete evidence that anisotropic geometry is an important but underexplored factor in IDE evaluation.
Overall, our results highlight a gap between standard synthetic benchmarks and more realistic geometric settings. \quest provides a controlled intermediate regime—retaining exact ground truth while exhibiting richer geometric structure—that can serve as a more informative testbed for future IDE methods.

\paragraph{Limitations} While \quest introduces controlled anisotropy and nontrivial geometry, it remains a synthetic benchmark and does not fully capture the complexity of real-world data distributions. Second, our analysis focuses on a limited set of estimators and computational regimes; a broader evaluation across more specialized methods may reveal additional insight.
Although we provide a scalable local covariance diagnostic for anisotropy, we do not claim that it is a complete differential-geometric invariant or that it fully explains all estimator failures. Establishing a quantitative causal model connecting anisotropy, curvature, \newtext{pullback metrics, second fundamental forms,} sampling scale, and IDE error remains an open line of investigation.
Finally, our experiments are limited by computational resources, restricting the scale of datasets and hyperparameter sweeps considered. These constraints may affect the generality of some quantitative conclusions. Additionally, we do not provide a quantitative correlation of anisotropy due to the lack of a suitable scalable measure available, to the best of our knowledge.

\paragraph{Future Work} A key direction for future work is to develop a more systematic understanding of geometric factors affecting intrinsic dimension estimation, beyond anisotropy alone. In particular, extending the benchmark to incorporate controlled variations in scale, noise structure, and “fuzzy” or non-manifold data could help bridge the gap to real-world datasets.
Another promising direction is the design of geometry-aware IDE methods that explicitly account for anisotropy in local neighborhoods.
\newtext{
With sufficient compute, we would extend Table~\ref{tab:perf} to a more diverse set of estimator families not covered here, including %
the scale-dependent Gride estimator \citep{DentiGride2022}, the geometry-aware GeoMLE estimator \citep{gomtsyan2019geometryawaremaximumlikelihoodestimation}, the normalizing-flow-based LIDL estimator \citep{tempczyk2021lidl}, and the diffusion-model-based FLIPD estimator \citep{kamkari2024a}. It would be interesting to see if (or how) anisotropy affects theese methods which do not perform explicit kNN computations.
Two more immediate open questions raised by our results are (i) formulating a general mechanistic interpretation of failure modes for IDE methods as investigated for lPCA and ABID. (ii) what property of the ``Gr (Vec)'' embedding specifically makes it robust to the anisotropy stressor that degrades the other four \quest embeddings. We leave these for future work.
}
Finally, expanding \quest to include a broader class of manifolds and integrating it with existing benchmarking libraries may facilitate more standardized and reproducible evaluation of intrinsic dimension estimators.

\section{Impact Statement}

This work contributes to the evaluation of intrinsic dimension estimation methods by identifying a previously underexplored failure mode—anisotropic geometry—and proposing a benchmark to study it in a controlled setting. Improved understanding of IDE behavior has implications for a range of applications in machine learning, including representation learning, data compression, and model complexity analysis.
As our benchmark is fully synthetic and does not involve human data, it raises no direct ethical or privacy concerns. However, more reliable intrinsic dimension estimation may influence how practitioners interpret the structure and complexity of real-world datasets, which could impact downstream modeling decisions.
We hope that this work encourages more rigorous and realistic benchmarking practices in the study of intrinsic dimension and related geometric properties of data.

\section*{Acknowledgements}
A.D.~thanks Andrey Gromov and Ishani Mondal for useful discussions, and acknowledges support from NSF DMR-2429246
J.T.I.~thanks the Joint Quantum Institute at the University of Maryland for support through a JQI fellowship.
V.V.A.~thanks Andrew Dienstfrey for useful comments.
We thank Erik Thorsden for providing code and access to ABID. 
The authors acknowledge the University of Maryland supercomputing resources (\url{http://hpcc.umd.edu}) made available for conducting the research reported in this paper.
Parts of this manuscript were prepared with the assistance of ChatGPT 5.6, developed by OpenAI, and Claude
Code, developed by Anthropic.  Certain products, commercial and otherwise, are mentioned in this publication. These mentions are for informational purposes only, and do not imply recommendation or endorsement by NIST.
Contributions to this work by NIST, an agency of the
US government, are not subject to US copyright.

\section*{Declaration of AI Usage}
The authors acknowledge the use of AI assistance from ChatGPT Codex and Claude Code, for the express purpose of formatting, grammatical checks, plotting and developing and maintaining the codebase.
All ideas presented in this work originate with the authors. All content, claims, and conclusions have been reviewed and verified by the authors to ensure accuracy and originality.

\bibliography{biblio}
\bibliographystyle{tmlr}

\appendix

\section{Licensing, Release, and Dataset Maintenance}
\label{app:lic}

The \quest dataset will be released under the \textbf{Creative Commons Attribution 4.0 International (CC BY 4.0)} license, allowing use, modification and redistribution with proper attribution. The dataset consists of embeddings for several quantum-inspired manifolds and contains no reference to any human data or sensitive information.

\textbf{Release and Maintenance plan: } The dataset will be released through a public GitHub repository. As presented in the Supplementary Material, scripts for generating embeddings of the different \quest manifolds will be uploaded, with clear instructions on the sampling process, including but not limited to relevant libraries, necessary computational budget, etc. 

Due to memory constraints, we will not be releasing the actual data samples, especially very high-dimensional ones. Instead, we will release the scripts used to generate samples from a particular \quest manifold embedding. 

The GitHub repository will also be tracked to allow users to report and fix bugs, and to propose additional manifold updates. We will further invite the community to contribute through bug reports and suggestions for new datasets and features. Each update will be accompanied by well-documented release notes that detail changes and new requirements. 

We anticipate integrating \quest with existing IDE libraries, providing additional avenues for long-term maintenance and community contributions. As such, since these are synthetic datasets, the original datasets can be maintained indefinitely. However, through the above practices, we hope to ensure that the dataset remains high-quality, well-maintained and easily accesible to the research community.

\textbf{Intended Use: } This dataset is intended primarily for research and benchmarking purposes. In particular, we envision these datasets to be useful and relevant for IDE performance evaluation and aspects of manifold hypothesis. 
Users are encouraged to cite the dataset and the accompanying paper when reporting results. While the dataset is released under an open license, any use must respect the intended research purpose and proper attribution requirements.

\textbf{Ethics and Privacy Standards: } No human subjects or sensitive information are involved in this dataset. The data is entirely synthetic, ensuring full compliance with privacy and ethical standards. Users are encouraged to adhere to best practices in computational research and reproducibility when using the dataset.

\section{Experimental details}
\label{app:exp-hprm}

We generated the data with the respective embeddings through custom scripts.

We tested 6 IDE methods --- lPCA, MLE, CorrInt, TwoNN, DANCo and ABID. Among these, lPCA was implemented manually, while ABID was implemented via the methodology shared by the authors of \citep{THORDSEN2022101989}. All other methods were obtained directly from the \texttt{scikit-dimension} package. All methods involved computing $k$ nearest neighbors (kNNs) which were pre-computed using the \texttt{Nearest Neighbors} module from sklearn.

For comparisons involving \quest and other benchmarks, we ran sweeps for the hyperparameter $k$ and the sample size $N$ for each of the methods. In particular, we performed three types of sweeps - sweeping $N$ logarithmically from 100 to 10000, holding $k$ fixed at $50$, sweeping $k$ logarithmically from 10 to 1000 while holding $N$ fixed at 5000 and sweeping $N$ from 100 to 10000 while holding $k/N$ fixed at the values [0.08, 0.1, 0.15, 0.2, 0.5, 0.99]. For DANCo, we ran a reduced hyper-parameter sweep, with $N \in [10, 200, 500, 1000, 2000] (k=10)$, $k \in [ 4, 5, 10] (N=2000)$ and $k/N \in [0.04,0.05,0.08, 0.1] with N \in [100,200,300]$. Any other hyper-parameters were held fixed at their default values, after small-scale experiments showed that the effect of these hyper-parameters were not as significant as compared to $k$ and $N$. All runs were performed with 3 different random seeds.

For the scaling experiments, we held $k$ fixed for all runs. The default values were $k = 100 (200)$ for lPCA; $k = 100$ for MLE, ABID, TwoNN; $k=10$ for DANCo; $k_1 = 10, k_2 = 20$ for CorrInt. In the case of $k >= N-1$, in which case we chose $k = N - 2$; for CorrInt, this was modified as $k_2 = N - 2$ and $k_1 = k_2 / 2$. We used $k = 100$ for the experiment where we scaled the sample size $N$, while we used $k = 200$ for the case of scaling with noise. All other hyper-parameters were kept at their default values. 

Experiments (including data generation) were run on dual AMD 7763 64-core CPUs and took around 40,000 CPU hours with an approximate total of 1000 hours for data generation. Plots and inferences were then made locally with negligible overhead cost. %

There are two dominant sources of errors for our plots --- spread in the local ID estimates for the entire sample, and the error due to using three different random seeds. The error reported is the mean error obtained for the local estimates, averaged over 3 experimental seeds.

\section{Detailed Background and Relevant Works}
\label{app:bg}

\paragraph{The manifold paradigm in ML} 
The manifold hypothesis has been a well-known paradigm in the machine learning community \citep{Gorban2018Manifold, Olah2014_MF, Cayton2005_MF}. This hypothesis has been subject to both theoretical and experimental investigations  \citep{fefferman2013testingmanifoldhypothesis, HiddenManifold, NIPS2010_8a1e808b_Manifold_exp, kiani2024hardness, brown2023verifyingunionmanifoldshypothesis, Donoho2005, Lee2003}. At the same time, there has been interest in understanding the topological structures of real-world datasets through the field of Topological Data Analysis \citep{Carlsson2008, Zia2024}. There is a separate notion involving manifolds in ML, which seeks to understand the learning process as modification of latent representations of the data manifold, as explored in \citep{magai2022topologygeometrydatamanifold, ansuini2019intrinsicdimensiondatarepresentations}.

\paragraph{Intrinsic Dimension} The intrinsic dimension (ID) of a manifold is a very useful quantity for several reasons. It is a characteristic property of the manifold, and hence many properties or features of learning problems are dependent on the ID --- dimensionality reduction \citep{Zhang2005, Gashler2011, Gashler2012}, exponential scaling of samples with ID \citep{Narayanan2009, NIPS2010_8a1e808b_Manifold_exp}, correlation of generalization capability with ID of data and internal representations \citep{pope2021the_MNIST_10, ansuini2019intrinsicdimensiondatarepresentations, birdal2021intrinsicdimensionpersistenthomology, magai2022topologygeometrydatamanifold, brown2022relatingregularizationgeneralizationintrinsic} as well as ID being a natural measure of local complexity of the data \citep{kamkari2024a}.

The importance of ID has thus led to a spate of research on developing estimators of ID. These methods take different approaches to estimating LID --- through various distance-based measures, such as those based on Euclidean distances \citep{1039212_CorrInt, Levina_MLE, gomtsyan2019geometryawaremaximumlikelihoodestimation, Facco_2017_TwoNN}, non-Euclidean geodetic distances \citep{Granata2016}, and Wasserstein distances \citep{block2022intrinsicdimensionestimationusing}; deviations of simplices \citep{Johnsson2014}; angle-based measures \citep{DANCOCERUTI20142569, THORDSEN2022101989}; measures based on generative models \citep{MNIST_100_stanczuk2023diffusionmodelsecretlyknows, tempczyk2021lidl, kamkari2024a, zheng2022learning}; and quantum encoding-based algorithms \citep{Candelori2025}. See \citep{CampadelliBench} for a more extensive survey of various ID estimators.

\paragraph{Benchmarks for IDE} \citep{CAMASTRA20032945} undertook one of the first surveys for ID estimators, outlining different datasets used for benchmarking ID estimators known at the time. \citep{Hein2005Benchmark} constructed a series of manifolds to benchmark their ID estimator, including hyperspheres, isotropic Gaussian vectors, the M{\"o}bius strip, etc. However, the authors acknowledged the problem with evaluating ID estimators for MNIST, and attempted to construct datasets with variable ID by applying transformations such as translation, rotation, etc. on MNIST images. \citep{Rozza2012} drew on these ideas and constructed several complicated manifolds, devising means of embedding data into higher dimensions without linear isometries.
\citep{CampadelliBench} took this further and proposed a benchmark with several such manifolds. A very interesting technique to benchmark ID estimators on distributions different from these manifolds was proposed in \citep{pope2021the_MNIST_10}. In their paper, the authors propose using GANs with certain restriction on the dimension $\bar{d}$ of the latent noise vectors to generate images, whose ID was then bounded from above by $\bar{d}$. However, this method suffers from an obvious problem in that there is no exact ground-truth ID for the generated data. A few other examples with modern IDE methods include the use of "lollipop", Swiss rolls, double helix, mutliscale manifolds in \citep{tempczyk2021lidl}, synthetic image manifolds in \citep{MNIST_100_stanczuk2023diffusionmodelsecretlyknows} and random affine transformations in \citep{kamkari2024a}. These benchmarks however are not simultaneously (i) scalable (ii) anisotropic or (iii) ID-preserving.

In addition, most ID estimators have a plethora of synthetic datasets on which they perform their zeroth-order evaluations - but most of them are low-dimensional, "simple" manifolds, whereas real-world data is more complicated and comes from vastly different distributions. There are some attempts to artificially induce complexity in these manifolds, examples of which are highlighted above. Scikit has compiled some of these manifolds into their python library \citep{Bac_2021Scikit}. However, it contains only a few topologically non-trivial manifolds such as the M{\"o}bius strip, Swiss roll, helixes, etc. which do not admit natural embeddings for arbitrary intrinsic dimensions. Previous attempts to construct physics-inspired datasets for IDE evaluation involved nonlinear dynamical systems. These include the Santa Fe datasets \citep{santafe_weigend1993time}, DSVC1 dataset \citep{camastra2009comparative} and strange attractors \citep{GRASSBERGER1983189_Corrint}. However, the true ID in these cases is not always known a priori, which makes it difficult to use as a benchmark. It is with this background that we propose \quest - an infinite family of manifolds which offer multiple natural notions of anisotropic embedding in higher-dimensional space. In addition, we propose a recipe to construct such manifolds through the concept of homogeneous spaces and coherent states. %
The key novelty of this benchmark is that it is able to induce a particular complexity, namely anisotropy, while maintaining control over other aspects like density, curvature, etc.

\section{Details for construction of \quest manifolds}
\label{app:proof}

We present a brief overview of the manifolds included in \quest. The interested reader is referred to the references herein for a more in-depth discussion.

All manifolds included in the \quest benchmark are parameterized by quotients \(\cal G/H\), where \(\cal G\) is a Lie group (i.e., a group that can also be considered as a manifold), and where \(\cal H\) is one of its subgroups. 
Such spaces are called homogeneous spaces, and they are examples of quotient spaces.
When \(\cal H\) is continuous subgroup, the intrinsic dimension of such quotient spaces is 
\(d_i = \dim {\cal G} - \dim {\cal H}\).
When \(\cal H\) is finite, $\dim \calH = 0$, and therefore the intrinsic dimension is equal to the dimension of \(\calG\).

Our embeddings are constructed using the Gilmore-Perelomov  method \citep{perelomov1977generalized-coh,perelomov1986generalized-coh,zhang1990coherent-states}, vectorization of matrices, and combinations thereof.
We review Gilmore-Perelomov coherent-states in Section~\ref{ssec:Gilmore-Perelomov} and use them to build the \quest manifolds in Section~\ref{ssec:eucldean-embeddings}.

\subsection{Gilmore-Perelomov embeddings}
\label{ssec:Gilmore-Perelomov}

Given the groups $\calH < \calG$ and some desired ambient dimension $d_A$ in which we wish to embed the manifold $\calM = \calG/\calH$,
we must first construct an orthogonal representation $\varLambda$ of $\calG$\footnote{\label{fn:embedding}When embedding into $\bbR^{d_A}$, the representation should be orthogonal; when embedding into $\bbC^{d_A}$, the representation should be unitary.}.
Define $\varLambda\vert_\calH$ to be the representation $\varLambda$ of $\calG$ restricted to $\calH$.
For compact Lie groups, all irreducible representations (irreps) appear in the isotypic decomposition of tensor-product representation \citep{58644}.
Our starting point will therefore be to consider representations of the form $\varLambda(g) = g^{\otimes t}$, where \(g\) is in the defining irrep, but the recipe can be generalized to other representations.

The projector onto the trivial irreps in the isotypic decomposition of a representation $\varLambda$ is $P(\varLambda) = \int_\calG \varLambda(g)~dg$, where $dg$ denotes the unique unit-normalized Haar measure on $\calG$.
We suppose that A). $P(\varLambda \rvert_\calH) > P(\varLambda)$, meaning that there are nontrivial irreps of \(\calH\) contained in \(\varLambda\), and B). that there are no subgroups $\calH < \calK < \calG$ with $P(\varLambda\rvert_\calK) > P(\varLambda\rvert_\calH)$.

Given the above assumptions, there exists at least one unit vector $v_{\calH}$ that lies in the image of $P(\varLambda\vert_\calH)$ and does not lie in the image of $P(\varLambda\rvert_\calK)$ for any $\calK > \calH$ (including $\calK = \calG$).
By construction, $v_\calH$ lies in a trivial irrep of $\varLambda\rvert_\calH$, and therefore $\varLambda(h)v_\calH = v_\calH$ for all $h\in \calH$.
Moreover, due to assumption (B), all other elements of \(\calG\) do not leave \(v_\calH\) invariant.
Thus, because any $g\in\calG$ can be uniquely written as $g = ah$ for a subgroup element $h\in\calH$ and coset representative $a \in \calM = \calG/\calH$, the mapping
\begin{equation}
    \psi_\calM(g) = \varLambda(g)v_\calH = \varLambda(a)\varLambda(h)v_\calH = \varLambda(a)v_\calH    
\end{equation}
is a well-defined, injective mapping between $\calM$ and $\bbR^{d_A}$ or $\bbC^{d_A}$ (see footnote \ref{fn:embedding}).
In fact, the embedding is injective into the real or complex sphere $\Omega_{d_A}$.

This mapping is equivariant -- $\varLambda(g_1)\psi_\calM(g_2) = \psi_\calM(g_1 g_2)$.
It is, however, not in general isometric -- this depends on the metric that is chosen on $\calM$. 
In particular, if the metric $g$ on $\calM$ is induced from the Cartan-Killing metric on $\calG$, then $\psi_\calM\colon (\calM, g) \to (\Omega_{d_A}, g_\Omega)$ will not be isometric, where $g_\Omega$ is the metric on the sphere.
Of course, we could instead choose $g$ to be the metric induced by the embedding; that is, the pullback $g = \psi_\calM^\ast g_\Omega$. In this case, the embedding is isometric by construction.

Finally, $\psi_\calM$ is an immersion simply because its derivative is everywhere injective.
Specifically, suppose that $A$ is a Lie algebra element so that $g(s) = g(0) e^{A s}$ is a curve in $\calM$. 
The embedded tangent space elements are then identified with 
$\frac{d}{ds}\psi_\calM(g(s))\rvert_{s=0}$.

\begin{wrapfigure}{r}{0.35\textwidth}
    \centering
    \includegraphics[width=0.35\textwidth]{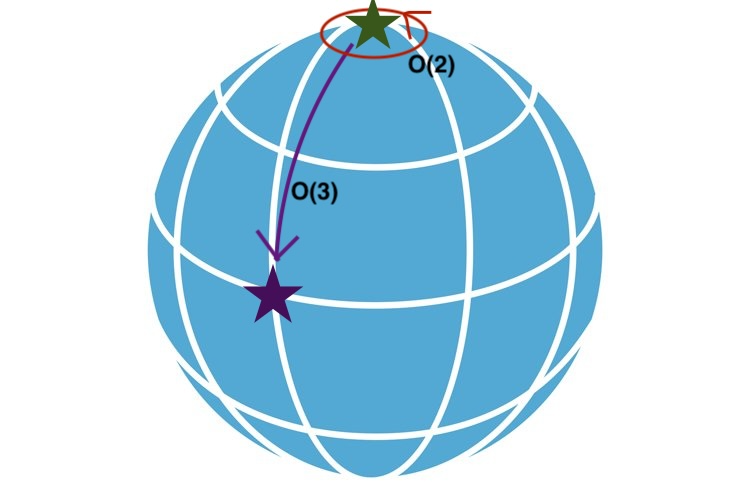}
    \caption{$2$-sphere is $\text{O}(3)/\text{O}(2)$.}
    \label{fig:2-sphere}
    \vspace{-20pt} %
\end{wrapfigure}

\subsubsection{Example of a Gilmore-Perelomov embedding}
\label{app:csm-sphere}

We use the method outlined above to parametrize points on the sphere:
\[
\mathbb{S}^2 \cong O(3)/O(2)
\]

\textbf{Step 1.} Pick a fixed vector: We choose the north pole which has the coordinates:
\[
v_o = \begin{bmatrix} 0 \\ 0 \\ 1 \end{bmatrix}
\]

\textbf{Step 2. Representations of the groups:}
\begin{itemize}
    \item A general rotation \( g \in SO(3) \) is parametrized by Euler angles:
    \[
    g(\alpha, \beta, \gamma) = R_z(\alpha) R_y(\beta) R_z(\gamma)
    \]
    where
    \[
    R_z(\theta) = 
    \begin{bmatrix}
        \cos\theta & -\sin\theta & 0 \\
        \sin\theta & \cos\theta & 0 \\
        0 & 0 & 1
    \end{bmatrix}, \quad
    R_y(\theta) =
    \begin{bmatrix}
        \cos\theta & 0 & \sin\theta \\
        0 & 1 & 0 \\
        -\sin\theta & 0 & \cos\theta
    \end{bmatrix}
    \]
    
    \item The stabilizer subgroup \( H = SO(2) \subset SO(3) \) consists of rotations about the \( z \)-axis:
    \[
    h(\gamma) = R_z(\gamma)
    \]

\end{itemize}

\textbf{Step 3. Invariance of the initial vector:}
\[
R_z(\gamma) v_0 = v_0 \qquad \Rightarrow \quad v_0 \text{ is invariant under } H
\]
Incidentally these are the only rotations which preserve the north pole.
\vspace{1em}

\textbf{Step 4. Construction of the embedding:}

\[
v_g = g(\alpha, \beta, \gamma) v_0
\]

Apply each part sequentially:
\[
R_z(\gamma)v_0 = v_0, \quad R_y(\beta)v_0 = 
\begin{bmatrix} \sin\beta \\ 0 \\ \cos\beta \end{bmatrix}, \quad
R_z(\alpha)
\begin{bmatrix} \sin\beta \\ 0 \\ \cos\beta \end{bmatrix}
=
\begin{bmatrix}
\sin\beta \cos\alpha \\
\sin\beta \sin\alpha \\
\cos\beta
\end{bmatrix}
\]

\textbf{Final result:}
\[
\boxed{
\psi_{\alpha, \beta} =
\begin{bmatrix}
\sin\beta \cos\alpha \\
\sin\beta \sin\alpha \\
\cos\beta
\end{bmatrix}
}
\]

This parametrizes the full sphere \( \mathbb{S}^2 \) in the usual polar coordinate form.

\subsubsection{Extension to natural data}

The notion of group representations and fiducial vectors can be easily extended to their discrete counterparts. If there exists a natural vectorization of the data, one can make the correspondence $\mathbf{x} \to  v_{\calH}$ by setting $\braket{e_i,v_\calH} = x_i$ where $e_i$ denotes some standard basis spanning a $d$-dimensional Hilbert space. One can then use these to generate an embedding for a homogeneous space as outlined above.

As an example, consider the number `1' in MNIST. The image admits translation invariance, and so one can use any standard image representing `1' as the fiducial vector to represent a group $\calG/\calT$ where $\calT < \calG$ represents discrete translations.

\subsection{Embedding manifolds into Euclidean space}
\label{ssec:eucldean-embeddings}

\begin{table}[h]
    \centering

\begin{tabular}{cccccc}
\toprule 
{\footnotesize \textbf{Family}} & {\footnotesize \textbf{Symbol}} & {\footnotesize \textbf{Quotient space}} & {\footnotesize$d_{i}$} & {\footnotesize \begin{tabular}[c]{@{}c@{}}\textbf{\quest}\\ \textbf{embedding}\end{tabular} } & {\footnotesize$d_{a}^{\textnormal{min}}$}\tabularnewline
\midrule
\multirow{2}{*}{{\footnotesize \begin{tabular}[c]{@{}c@{}}Stiefel\\ manifold\end{tabular} }} & \multirow{2}{*}{{\footnotesize$\ensuremath{\St(k,\bbR^{n})}$}} & \multirow{2}{*}{{\footnotesize$\frac{\O(n)}{\O(n-k)}$}} & \multirow{2}{*}{{\footnotesize${\scriptstyle nk-\frac{1}{2}k(k+1)}$}} & {\footnotesize St (Matrix)} & {\footnotesize${\scriptstyle nk}$}\tabularnewline
\cmidrule{5-6}
 &  &  &  & {\footnotesize St (Vec)} & {\footnotesize${\scriptstyle \binom{n}{k}+k^{2}}$}\tabularnewline
\midrule 
\multirow{2}{*}{{\footnotesize \!\!\!\! Grassmannian}} & {\footnotesize$\ensuremath{\Gr(k,\bbR^{n})}$} & {\footnotesize$\frac{\O(n)}{\O(n-k)\O(k)}$} & {\footnotesize${\scriptstyle k(n-k)}$} & {\footnotesize Gr (Proj)} & {\footnotesize${\scriptstyle n^{2}}$}\tabularnewline
\cmidrule{5-6}
 & {\footnotesize$\ensuremath{\Gr^{\star}(k,\bbR^{n})}$} & {\footnotesize$\frac{\O(n)}{\O(n-k)\SO(k)}$} & {\footnotesize${\scriptstyle k(n-k)}$}  & {\footnotesize Gr (Vec)} & {\footnotesize${\scriptstyle \binom{n}{k}}$}\tabularnewline
\midrule 
{\footnotesize \begin{tabular}[c]{@{}c@{}}Flag\\ manifold\end{tabular} } & {\footnotesize Flag$^{\star}(k_{1},k_{2},\bbR^{n})$} & {\footnotesize$\frac{\O(n)}{\SO(k_{1})\SO(k_{2})\O(n-k_{1}-k_{2})}$} & {\footnotesize$\substack{(k_{1}+k_{2})n\\
-k_{1}^{2}-k_{2}^{2}-k_{1}k_{2}
}
$} & {\footnotesize Flag (Vec)} & {\footnotesize$\binom{n}{k_{1}}\binom{n}{k_{2}}$}\tabularnewline
\bottomrule
\end{tabular}
\vspace{0.07in}

    \caption{
    Table listing manifold families used in the \quest benchmark, their mathematical symbols, their equivalent quotient/homogeneous spaces, and their intrinsic dimensions \(d_i\). 
    Here, \(\O(n)\) denotes the orthogonal group in \(n\) dimensions, and \(\SO(n)\) denotes the special orthogonal groups.
    The parameter \(n \geq 1\).
    The last two columns list the five \quest embeddings and their lowest possible ambient dimensions, \(d_a^{\text{min}}\).
    Any embedding into a given ambient dimension can be further embedded into a space of larger ambient dimension via any isometry.
    The use of slightly different quotients for our two Grassmannian embeddings yields a lower possible ambient dimension for the latter embedding while maintaining the same intrinsic dimension.
    Note that the usual flag manifold can be obtained from Flag\(^{\star}(k_{1},k_{2},\bbR^{n})\) by letting \(\SO\to\O\).
    The symbol \(\binom{a}{b}\) is the binomial coefficient.
    }
    \label{tab:manifold-params}
\end{table}

Throughout this section, let \( \set{e_1,\dots,e_n} \) be the standard (unit-vector) basis of \( \bbR^n \).
Furthermore, for a matrix \( X \), define \( \vec X \) to be the vectorization of \( X \) -- that is, the vector obtained by stacking the columns.

\subsubsection*{Stiefel manifolds and the ``St (Matrix)'' embedding}

The Stiefel manifold is \( \St(k, \bbR^n) = \set{X \in \bbR^{n\times k} \mid X^T X = \bbI} \) \citep{james1977the-topology-of,bendokat2020a-grassmann-man}.
This easily embeds into \( \bbR^{nk} \) via the map \( X \mapsto \vec X \).
This yields the ``St (Matrix)'' embedding in the \quest dataset.

A more interesting embedding will be used as an intermediate step to derive the Grassmannian embedding used in \quest and is as follows.
We will view \( \St(k, \bbR^n) \) as \( \O(n)/\O(n-k) \).
Then a point \( X \in \St(k, \bbR^n) \) is given by the first \( k \) columns of the corresponding orthogonal matrix.
Using the notation from Section~\ref{ssec:Gilmore-Perelomov},
we let $\calG = \O(n)$ and $\calH = \O(n-k)$, and
define the representation \( \varLambda\colon \calO \mapsto \calO^{\otimes k} \) for \( \calO\in \O(n) \).
An obvious choice for the state $v_\calH$ is $e_1\otimes \dots \otimes e_k$.
Given \( \calO\in\O(n) \), this yields the embedding
\begin{align}
    \psi_\St(\calO)
    &= \varLambda(\calO) v_\calH \\
    &=  \calO e_1 \otimes \calO e_2 \otimes \dots \otimes \calO e_k \\
    &= \sum_{j_1,\dots,j_k=1}^n \calO_{j_1,1}\dots \calO_{j_k,k} e_{j_1}\otimes \dots\otimes e_{j_k} \\
    &= \calO_{\ast,1}\otimes \dots \otimes \calO_{\ast,k},
\end{align}
where \( \calO_{\ast,i} \) denotes the \( i^{\rm th} \) column of \( \calO \).
From this we can easily see that \( \psi_\St(\calO) \) depends only on the first \( k \) columns of \( \calO \), but will result in a different vector for different elements of the Stiefel manifold.
It is therefore an injection of the Stiefel manifold into \( \bbR^{n^k}\).

\subsubsection*{Grassmannians and the ``Gr (Vec)'' embedding}

Consider the embedding \( \psi_\St(\calO) \) of the Stiefel manifold from above. 
We need to slightly modify this to get an embedding of the Grassmannian,
\begin{equation}
    \label{eq:grassmannian}
   \ensuremath{\Gr^{\star}(k,\bbR^{n})}\cong\frac{\St(k,\bbR^{n})}{\SO(k)}\cong\frac{\O(n)}{\O(n-k)\times \SO(k)}\,,   
\end{equation}
due to the extra quotient by \( \SO(k) \).
In particular, we define 
\begin{equation}
    v_\calH = 
    \frac{1}{\sqrt{k!}} \sum_{\sigma\in S_k} \text{sgn}(\sigma) 
    e_{\sigma(1)}\otimes \dots\otimes e_{\sigma(k)} ,
\end{equation}
where $\text{sgn}(\sigma)$ denotes the symmetric group on $k$ elements.

We note that Eq.~\ref{eq:grassmannian} is not the usual definition of the Grassmannian.
Indeed, Grassmannian is typically $\Gr(k, \bbR^n) = \frac{\O(n)}{\O(n-k)\times \O(k)}$, and the oriented Grassmannian is $\widetilde\Gr(k, \bbR^n) = \frac{\SO(n)}{\SO(n-k)\times \SO(k)}$.
These two spaces, along with the space in Eq.~\ref{eq:grassmannian}, all have the same ID.
In future work, we will add the standard and oriented Grassmannian to \quest and further compare the performance of standard ID estimation methods on these three very similar manifolds.
In particular, the geometry and topology of these three manifolds are slightly different despite the manifolds themselves being morally very similar, thus yielding an interesting testing ground.
For example, we list the first two homotopy groups of these manifolds in Table.~\ref{tab:homotopy}, which we calulate using the long exact sequence of homotopy groups induced from the fiber bundle $\calH \to \calG \to \calG / \calH$.

\begin{table}[]
    \centering
    \begin{tabular}{rlcc}
    \toprule
        \( \calM \quad= \)                   & \( \calG/\calH \)           & \( \pi_1(\calM) \) & \( \pi_2(\calM) \) \\
        \midrule
        $\Gr(k, \bbR^n) \quad=$ & \( \O(n) / \O(n-k)\times \O(k) \)    & \( \bbZ_2 \)   & \( \bbZ_2 \)   \\
        $\widetilde\Gr(k, \bbR^n) \quad=$ & \( \SO(n) / \SO(n-k)\times \SO(k) \) & \( 0 \)        & \( \bbZ_2 \)   \\
        $\Gr^\star(k, \bbR^n) \quad=$ & \( \O(n) / \O(n-k)\times \SO(k) \)   & \( \bbZ_2 \)   & \( \bbZ_2 \) \\
        \bottomrule
    \end{tabular}
    \caption{The first two homotopy groups of the three types of ``Grassmannians'' described below Eq.~\ref{eq:grassmannian}.}
    \label{tab:homotopy}
\end{table}

Given $\calH$, we achieve the embedding
\begin{align}
    \psi_\Gr(\calO)
    &= \varLambda(\calO)v_\calH \\
    &= \frac{1}{\sqrt{k!}} \sum_{\sigma\in S_k} \text{sgn}(\sigma) \calO e_{\sigma(1)}\otimes \dots\otimes \calO e_{\sigma(k)} \\
    &= \frac{1}{\sqrt{k!}} \sum_{\sigma\in S_k}\sum_{j_1,\dots,j_k=1}^n  \text{sgn}(\sigma) \calO_{j_1,\sigma(1)}\dots \calO_{j_k,\sigma(k)}  e_{j_1}\otimes \dots \otimes e_{j_k} \\
    &= \frac{1}{\sqrt{k!}} \sum_{j_1,\dots,j_k=1}^n
    \det(\calO_{\parentheses{j_1,\dots,j_k}})
    e_{j_1}\otimes \dots \otimes e_{j_k} ,
\end{align}
where \( \calO_{\parentheses{j_1,\dots,j_k}} \) denotes the \( k\times k \) matrix obtained from \( \calO \) by taking the first \( k \) columns and taking the rows \( j_1,\dots,j_k \),
and similarly
\( \calO_{\set{j_1,\dots,j_k}} = \calO_{\operatorname{sorted}\parentheses{j_1,\dots,j_k}} \).

Let \( \sigma\in S_k \) be the permutation that sorts \(  \parentheses{j_1,\dots,j_k} \).
Then \( \det \calO_{\parentheses{j_1,\dots,j_k}} = \text{sgn}(\sigma) \det \calO_{\set{j_1,\dots,j_k}} \).
Therefore, we have that
\begin{align}
    \psi_\Gr(\calO)
    &= \sum_{\substack{Q \subset \set{1,\dots, n} \\ \abs{Q} = k}} \det \calO_Q
    \frac{1}{\sqrt{k!}}\sum_{\sigma\in S_k} \text{sgn}(\sigma)
    e_{\sigma(Q_1)}\otimes \dots \otimes e_{\sigma(Q_k)} \\
    &= \sum_{\substack{Q \subset \set{1,\dots, n} \\ \abs{Q} = k}} \det \calO_Q
    u_Q,
\end{align}
where we defined \( Q_i \) to be the \( i^{\rm th} \) element of the set \( Q \) (of course this does not make sense generally, but because we are summing over all permutations, it is fine to pick some arbitrary ordering of the set),
and we defined the state
\begin{equation}
    u_Q = \frac{1}{\sqrt{k!}}\sum_{\sigma\in S_k}\text{sgn}(\sigma)
e_{\sigma(Q_1)}\otimes \dots \otimes e_{\sigma(Q_k)}~.
\end{equation}

Notice that we are embedding into \( \bbR^{n^k} \), but a full basis for the space is given by \( u_Q \) for all \( Q\subset \set{1,\dots, n} \) with \( \abs{Q} = k \).
Thus, this embedding in general gives us an embedding into \( \bbR^{d_A} \) for any \( d_A \geq \binom{n}{k} \).
We denote this by the shorthand ``Gr (Vec)'' in Table~\ref{tab:manifold-params}.

We note that this embedding is almost the Pl{\"u}cker embedding \citep{bendokat2020a-grassmann-man}.
The Pl{\"u}cker embedding is an embedding of the standard Grassmannian into real projective space, which is the real sphere with antipodal points identified.
Above, we are embedding Eq.~\ref{eq:grassmannian} into the real sphere.
The reason that $\psi_\Gr$ is not an embedding of $\O(n) / \O(n-k)\times \O(k)$ is because the vector $v_\calH$ is not invariant under the action $\O(k)$. 
Instead, it is invariant under the action of $\SO(k)$, and yields a \(\pm 1\) phase factor under the action of $\O(k)$.
This yields a well-defined embedding of $\O(n) / \O(n-k)\times \O(k)$ into projective space, but further embedding projective space into the sphere via the standard maps (e.g., $ (x_i) \mapsto (x_i x_j)_{i \leq j} $ for projective-space vectors \((\cdots x_j \cdots )\)) would come at a price of quadratically increasing the lowest possible ambient dimension \citep{simanca2018canonicalisometricembeddingsprojective,4097673,3505749}.
We thus stick with our original mapping in order to be able to run smaller-scale numerical experiments.

\subsubsection*{Stiefel manifold revisited: the ``St (Vec)'' embedding}

A point in the Stiefel manifold can be represented by a point on the Grassmannian and a matrix \( \mathcal{V} \in \SO(k) \).
In other words, any element of \( \O(n)/\O(n-k) \) can be expressed as an element of \( \O(n)/\O(n-k)\times\SO(k) \) and an element of \( \SO(k) \).
Therefore, given an element \( (\calO, \mathcal{V}) \) on \( \St(k,\bbR^n) \), where \( \calO\in\O(n) \) represents a point on \( \Gr^{\star}(k, \bbR^n) \) and \( \mathcal{V} \in \SO(k) \), we can define an embedding
\begin{equation}
    \tilde\psi_\St = \psi_\Gr(\calO) \oplus \vec{\mathcal{V}}~,
\end{equation}
recalling that \(\vec {\mathcal{V}}\) is the vectorized matrix \(\mathcal{V}\).
This embedding has dimension \( d + k^2 \), where \( d\geq \binom{n}{k} \), and we denote this by ``St (Vec)''.
This is much less than the dimension of the embedding \( \psi_\St \), which is \( n^k \).

In the numerical implementation, we independently sample
\(\calO\in\O(n)\) and \(\mathcal{V}\in\O(k)\), and apply the same direct-sum formula.
The \(\O(k)\) factor has two isometric determinant components, each with the same local intrinsic dimension as \(\SO(k)\).

\subsubsection*{The ``Gr (Proj)'' embedding}

Recall that $\Gr(k,\bbR^n)=\O(n)/\O(n-k)\times\O(k)$ is the manifold of $k$-dimensional subspaces of $\bbR^n$.
Thus, we can uniquely represent a point on this manifold by a projector that projects onto this corresponding subspace.
In particular, given an $n\times k$ orthogonal matrix $\calO, \calO^T \calO = \bbI_{k\times k}$ representing a point on $\St(k, \bbR^n)$, we can create the projector $P_\calO = \calO \calO^T$ that projects onto the span of the columns of $\calO$.
From this projector, we define the ``Gr (Proj)'' embedding as its vectorization $\vec P_\calO$.

As is, the Gr (Vec) and Gr (Proj) embed different spaces --- $\O(n)/\O(n-k)\times\SO(k)$ versus $\O(n)/\O(n-k)\times\O(k)$ --- but the intrinsic dimension remains the same.

\subsubsection*{Flag manifolds and the ``Flag (Vec)'' embedding}

In this section, we consider a general flag manifold \( \frac{\O(n)}{\SO(k_1)\times \dots \times \SO(k_t) \times \O(n-k)} \) where \( \sum_{i=1}^t k_i = k \).
We note, as with the Grassmannians, that our definition of the Flag manifolds also slightly differs from the standard. Namely, 
the typical Flag manifold is 
\( \frac{\O(n)}{\O(k_1)\times \dots \times \O(k_t) \times \O(n-k)} \).
We will add these to \quest in future work.

We begin with the case \(t=2\), as presented in Table~\ref{tab:manifold-params}. 
Again, in the notation of Section~\ref{ssec:Gilmore-Perelomov}, we work with the representation $\varLambda\colon \calO \mapsto \calO^k$, and we use the fiducial vector 
\begin{equation}
    v_\calH
    =
    \parentheses{\frac{1}{\sqrt{k_{1}!}}\sum_{\sigma\in S_{k_{1}}}e_{\sigma(1)}\otimes\dots\otimes e_{\sigma(k_{1})}}
    \otimes 
    \parentheses{\frac{1}{\sqrt{k_{2}!}}\sum_{\sigma\in S_{k_{2}}}e_{\sigma(k_{1}+1)}\otimes\dots\otimes e_{\sigma(k_{1}+k_{2})}}
    .
\end{equation}
This yields the embedding
\begin{align}
\psi_{F}(\calO)
&= \varLambda(\calO) v_\calH\\
&={\scriptscriptstyle \frac{1}{\sqrt{k_{1}!k_{2}!}}}\sum_{j_{1},\dots,j_{k}=1}^{n}{\scriptstyle \det\pargs{\calO_{(j_{1},\dots,j_{k_{1}}),(1,\dots,k_{1})}}\det\pargs{\calO_{(j_{k_{1}+1},\dots,j_{k}),(k_{1}+1,\dots,k)}}e_{j_{1}}\otimes\dots\otimes e_{j_{k}}}\\&={\scriptstyle \frac{1}{\sqrt{k_{1}!k_{2}!}}}\sum_{\substack{Q\subset\set{1,\dots,n}\\
\abs{Q}=k_{1}
}
}\sum_{\substack{P\subset\set{1,\dots,n}\\
\abs{P}=k_{2}
}
}\sum_{\sigma\in S_{k_{1}}}\sum_{\pi\in S_{k_{2}}}\times\\&{\scriptstyle \quad\times\text{sgn}(\sigma)\text{sgn}(\pi)\det\pargs{\calO_{Q,(1,\dots,k_{1})}}\det\pargs{\calO_{P,(k_{1}+1,\dots,k)}}e_{\sigma(Q_{1})}\otimes\dots\otimes e_{\sigma(Q_{k_{1}})}\otimes e_{\pi(P_{1})}\otimes\dots\otimes e_{\pi(P_{k_{2}})}}\nonumber\\&=\sum_{\substack{Q\subset\set{1,\dots,n}\\
\abs{Q}=k_{1}
}
}\sum_{\substack{P\subset\set{1,\dots,n}\\
\abs{P}=k_{2}
}
}\det\pargs{\calO_{Q,\set{1,\dots,k_{1}}}}\det\pargs{\calO_{P,\set{k_{1}+1,\dots,k}}}u_Q\otimes u_P.
\end{align}

The numerical implementation uses nested column prefixes.
Writing their ranks as \(r\geq s\), for \(r>s\), Hodge duality relates this construction to the disjoint-block construction above with block sizes \((n-r,s)\), up to a fixed orthogonal change of coordinates, with the same sampling law.
For equal prefix ranks \(r=s\), the image has intrinsic dimension \(r(n-r)\).
We call this the ``Flag (Vec)'' embedding.

A straightforward extension to general \(t\) yields an embedding of \( \frac{\O(n)}{\SO(k_1)\times \dots \times \SO(k_t) \times \O(n-k)} \) into \( \bbR^d \) for any ambient dimension \( d \geq \prod_{i=1}^t \binom{n}{k_i} \).

\subsection{Extension to double coset spaces}

The Gilmore-Perelomov embedding, introduced above, produces vectors \(\psi_{{\cal G}/{\cal H}}(a)\) valued in a homogeneous space \(\mathcal{M} = \mathcal{G}/\mathcal{H}\) for coset representatives \(a\).
Such vectors transform under elements \(g \in \cal G\) according to the group's multiplication rules, \(|\psi_{{\cal G}/{\cal H}}(a)\rangle \to |\psi_{{\cal G}/{\cal H}}(ga)\rangle\).

A double coset space \(\mathcal{K}\backslash \mathcal{G}/\mathcal{H}\) can be constructed from these vectors by taking their orbits under the subgroup \(\cal K\),
\begin{equation}
    v_{\mathcal{K}a\mathcal{H}}=\frac{1}{\sqrt{|\calK|}}\sum_{k\in{\cal K}} \psi_{{\cal G}/{\cal H}}(ka)
\end{equation}
for all coset representatives \(a\). Above, we use a sum to denote the linear combination of vectors, but this can substituted for an integral over the Haar measure in the case of a continuous group.

As an example, let us take the case of the sphere, \(\calM = \text{SO}(3)/\text{SO}(2)\), and consider \(\calK = \text{SO}(2)\) to be the group of proper rotations around the \(\mathbf{z}\)-axis.
The orbit of a generic point on the sphere under this subgroup is the latitude on which the point is located.
Therefore, double cosets are labeled by points in a fixed longitude connecting the north and south poles, which is no longer a manifold but an orbifold \citep{albert2020robust}.
The reason is because the poles do not move under \(\cal{K}\)-rotations and are singular points whose neighborhood does not resemble flat space.

Double-coset spaces for finite subgroups of \(\text{SO}(2)\) come with enlarged versions of a longitudinal line that are called spherical lunes. In such cases, the poles remain the two singular points of these orbifolds. In general, orbifolds of other discrete subgroups of \(\text{SO}(3)\) correspond to fundamental domains --- patches of the sphere whose translates under \(\calK\) tile the entire sphere.

\section{Overview of IDE methods tested}
\label{app:std-skdim}

\paragraph{lPCA :} Since a manifold is locally isomorphic to $\mR^{d_i}$, the directions normal to the hyperplanes have zero variance.  Given $k$ nearest-neighbor (NN) of a point sampled from $\mathbb{R}^D$, singular value decomposition (SVD) of the $k \times D$ matrix is performed to determine the principal components. The intrinsic dimension of the manifold can be then estimated by different means: 
\begin{itemize}
    \item 'maxgap' : the component showing the biggest spectral gap is returned as the intrinsic dimension, i.e.
    \begin{equation}
        \label{eq:pca-mg}
        \hat{d_i}(x;k) = \text{argmax}_{j\in 1, \dots,\min(D,k)-1} \frac{e_{j-1}}{e_{j}}
    \end{equation}
    \item 'ratio' : the minimum number of components needed to explain $1 - \epsilon$ of the total variance.
    \begin{equation}
        \label{eq:pca-ratio}
        \hat{d_i}(x;k, \epsilon) = \min \{ j : R_j \geq 1 - \epsilon \}
    \end{equation}
    where $R_j := \frac{\sum_{i=1}^j \sigma^2_{[i]}}{\sum_{i=1}^N \sigma^2_{[i]}}$ is the cumulative variance ratio ($(.)_{[i]}$ denotes that the quantity is sorted from largest to smallest (algebriacally).
    \item 'fo' : the index $j$ for which the (sorted) eigenvalues cross $(1-\epsilon)$ of the largest eigenvalue.
    \begin{equation}
        \label{eq:pca-fo}
        \hat{d_i}(x;k, \epsilon) = \min \{ j : \sigma^2_{[i]} \geq (1 - \epsilon) \sigma^2_{[k]} \}
    \end{equation}
\end{itemize}
It is easy to check that these definitions are equivalent with suitable choices of the hyperparameter $\epsilon$ for manifolds with a \textit{clear} spectral gap. In order to preserve computational resources, we
therefore report the results of the 'maxgap' technique in the main section. We also checked results with the 'ratio' and 'fo' versions in several small-scale experiments. We noticed that in order to match the results, careful hyper-parameter tuning needs to be done for the other methods. As a result, we omit discussing these results here.

We created a custom function to compute the $\hat{d}_i$ according to Eq. \ref{eq:pca-mg} above, closely drawing from the implementation in \href{https://scikit-dimension.readthedocs.io/en/latest/skdim.id.lPCA.html}{scikit-dimension}.

The following methods, namely \href{https://scikit-dimension.readthedocs.io/en/stable/skdim.id.MLE.html}{MLE}, \href{https://scikit-dimension.readthedocs.io/en/latest/skdim.id.DANCo.html}{DANCo}, \href{https://scikit-dimension.readthedocs.io/en/latest/skdim.id.CorrInt.html}{CorrInt} and \href{https://scikit-dimension.readthedocs.io/en/latest/skdim.id.TwoNN.html}{TwoNN} were implemented by accessing the implementations directly from the \texttt{scikit-dimension} package.

\paragraph{MLE : }  Given a set of samples $X_1, X_2, ... X_n$, related to a sample in lower-dimensional space $m$ and equipped with a smooth density $f$, one considers the Poisson process $\lambda(t)$ of the number of points in a small sphere around the point with radius $t$. The log-likelihood of this process (assuming a constant density $f(x)$ in the sphere of radius $R \geq t$) yields 

\begin{equation}
    \label{eq:mle-lev}
    \hat{d_i}(x; R) = \left[ \frac{1}{N_R(x)} \sum_{j=1}^{N_R(x)} \log \frac{R}{T_j(x)} \right]^{-1}
\end{equation}

where $T_j(x)$ is the Euclidean distance between $x$ and its $j$-th NN. Some results on the statistics of these log distances were considered too. 

\citep{Haro2008MLE} goes one step further and considers a noisy translation of the observed distances, which leads to a non-linear recursive equation, which they can solve self-consistently. In particular for isotropic Gaussian noise with scale $\sigma$, their ID estimate reads

\begin{equation}
    \label{eq:mle-haro}
    \hat{d_i}(x; R) = \left[ \frac{1}{N_R(x)} \sum_{j=1}^{N_R(x)} \frac{\int_{r=0}^R \exp(-\frac{(T_i- r)^2}{2 \sigma^2}) \log (\frac{T_k}{r}) dr }{\int_{r=0}^R \exp(-\frac{(T_i- r)^2}{2 \sigma^2}) dr} \right]^{-1}
\end{equation}

\paragraph{CorrInt : }  For a set $X_1, . . . , X_n$ of i.i.d. samples with a smooth density $f(x)$ in $\mathbb{R}^{d_i}$, the Euclidean distance between a point $x$ and its $k$-th NN $T_k(x)$ satisfies
\begin{equation}
    k/n \approx f(x) V_{d_i}(T_k(x))
\end{equation} where $V_{d_i}(R)$ is the volume of a $d_i$-dimensional sphere of radius $R$. Since $V_{d_i}(R) \propto R^{d_i}$, the number of points $k$ in a ball of radius $R$ grows exponentially with $d_i$. This is the motivation behind the fractal dimension definition.

One computes the correlation integral (sum) as

\begin{equation}
    \label{eq:corr-int-def}
    C(r) = \frac{1}{N(N-1)} \sum_{i > j} \mathbf{1}_{|| \mathbf{x}_i - \mathbf{x}_j || \leq r}
\end{equation}

One can then estimate the dimension as
\begin{equation}
    \hat{d_i}(x; k_1, k_2)= \frac{\log(C(r_2)/C(r_1))}{\log(r_2/r_1)}
\end{equation}

where the hyper-parameters $k_1$ and $k_2$ are used to find the median distances $r_1$ and $r_2$

\paragraph{TwoNN : } For a constant density $\rho$ around a point $x$, the volume of the hyper-spherical shell between $i$ and $i+1$-th NN is drawn from an exponential distribution in the volume $\Delta \nu_l = \omega_{d_i}(r^{d_i}_l - r^{d_i}_{l-1})$. Define $R = \frac{\Delta \nu_2}{\Delta \nu_1}$ and then it follows that $f(R) = (1+R)^{-2}$. It then follows that $f(\mu) = {d_i}\mu^{-{d_i}-1}$ where $\mu = \frac{r_2}{r_1} \in [1, \inf )$. (Basically note that $f(\mu) = f(R) \frac{dR}{d\mu} $ and $R = \mu^{d_i} - 1$)  To make matters less prone to computational errors, discard $\alpha$ of the largest $\frac{r_2}{r_1}$ values. Thus the ID estimate is given by
\begin{equation}
    \label{eq:id-twonn}
    \hat{d}_i(x; k, \alpha) = -\frac{\log(1 - F(\mu))}{\log \mu}
\end{equation}
where $\mu$ represents the ratio $\frac{r_2}{r_1}$ for the $k$NNs of the particular point.

\paragraph{DANCo : }
For a manifold $\calM \subseteq \mR^{d_i} $, consider an embedding $\phi : \mR^{d_i} \to \mR^D$ which is locally isometric, smooth and possibly non-linear. Then the points in a local neighborhood are drawn uniformly from the hyperspheres. The distribution for distances for the unit hypersphere normalized by the distance of the $k$-th NN follows the distribution

$$ g(r; k, d_i) = k d_i r^{d_i - 1} (1 - r^{d_i})^{k-1}$$

while the mutual angles follow the von Mises-Fisher (VMF) distribution 

$$ q(\mathbf{x}; \bf{\nu}, \tau) = C_{d_i}(\tau) \exp( \tau \bf{\nu}^T \mathbf{x}) $$ 

where $C_{d_i}(\tau)$ is a normalization constant. It should be noted that the parameter $\tau$ in the VMF distribution denotes the concentration of angles around the mean --- the parameter $\tau = 0$ reducing this distribution to the uniform distribution on the sphere. The joint distribution of the normalized distance and mutual angles factorizes into the product of marginals for the unit hypersphere. The ID is then estimated by minimizing the KL-divergence between the theoretical and experimental joint distribution of normalized distance and mutual angles.

\begin{equation}
    \label{eq:id-danco}
    \hat{d}_i(x ; k) = \text{argmin}_{d = 1, \dots, d_a} \int_{-\pi}^{\pi} d\theta \int_0^1 dr  h_d(r, \theta) \log( \frac{\hat{h}_d(r, \theta)}{h_d(r, \theta)})
\end{equation}

where $\hat{h}$ refers to the experimental joint distribution, and $h(r, \theta) = g(r) \cdot q( \theta)$ is the theoretical joint distribution.

\paragraph{ABID : }  As discussed in \citep{THORDSEN2022101989}, the distribution of pairwise cosines between two points drawn randomly and uniformly from a $d$-ball (excluding the origin; also holds for any such spherical distribution) follows a Beta distribution on the interval [-1,1].

$$ P(\cos \theta) = \frac{1}{2} B(\frac{1 + \cos \theta}{2}; \frac{d_i - 1}{2}, \frac{d_i - 1}{2}) $$

from which it follows that 

$$ \mathbb{E}[ \cos^2 \theta] = d_i^{-1}$$

This motivates the following definition for the ABID ID estimator

\begin{equation}
    \label{eq:id-abid}
    \hat{d}_i (x ; k) = (\mathbb{E}_{\mathbf{x}_i,\mathbf{x}_j \sim B_k(x)}[ \cos^2(\mathbf{x}_i, \mathbf{x}_j) ])^{-1}
\end{equation}

where $B_k(x)$ denotes the ball containing the kNN of x, i.e. a ball of radius $T_k(x)$

\section{Additional Experiments}

\subsection{Local anisotropy diagnostics}
\label{app:anisotropy}

We use a simple sample-based diagnostic for local anisotropy. For each anchor point $x_i$, we compute its $k$ nearest neighbors in the ambient Euclidean metric, add the anchor point to this neighborhood, center the resulting neighborhood, and form the local covariance matrix
\begin{equation}
    C_i^{(k)}
    =
    \frac{1}{m-1}
    \left(Y_i^{(k)}\right)^\top Y_i^{(k)} ,
\end{equation}
where $Y_i^{(k)}$ is the centered neighborhood matrix and $m=k+1$. Equivalently, if $Y_i^{(k)}=U\Sigma V^\top$, then the covariance eigenvalues are $\lambda_j=\sigma_j^2/(m-1)$. We normalize these eigenvalues by their sum and compare the resulting spectra across anchor points.

\newtext{
Under invariant sampling, a symmetry-based argument shows that the local covariance spectra should have a staircase of flat plateaus in the limit \(k,N\to\infty\), \(k/N\to 0\).
If a symmetric matrix $C$ commutes with a group of rotations, then its eigenspaces are invariant under the group: for any eigenvector $v$ with $Cv=\lambda v$ and any rotation $R$, we have
$$C(Rv)=R(Cv)=\lambda Rv.$$
Hence, on every irreducible invariant subspace of the rotation-group representation, $C$ must act as a scalar multiple of the identity by Schur's lemma. Therefore, all eigenvalues within each such block are equal.
Applying this to the population local covariance at an anchor point $p$ with rotations that map the local manifold to itself while keeping $p$ fixed,
we observe that the expected covariance is invariant. \footnote{This is because such rotations have the effect of permuting samples and orthogonal transformations only, which leave the singular values invariant.} As a consequence, the population local covariance
spectrum of a \quest manifold is a staircase of exactly flat plateaus. For a sphere, the top plateau (the tangent block) has multiplicity exactly $d_i$ and is exactly flat at
every scale, not just asymptotically.
}

\begin{figure}[t]
    \centering
    \begin{subfigure}{0.49\textwidth}
        \centering
        \includegraphics[width=\textwidth]{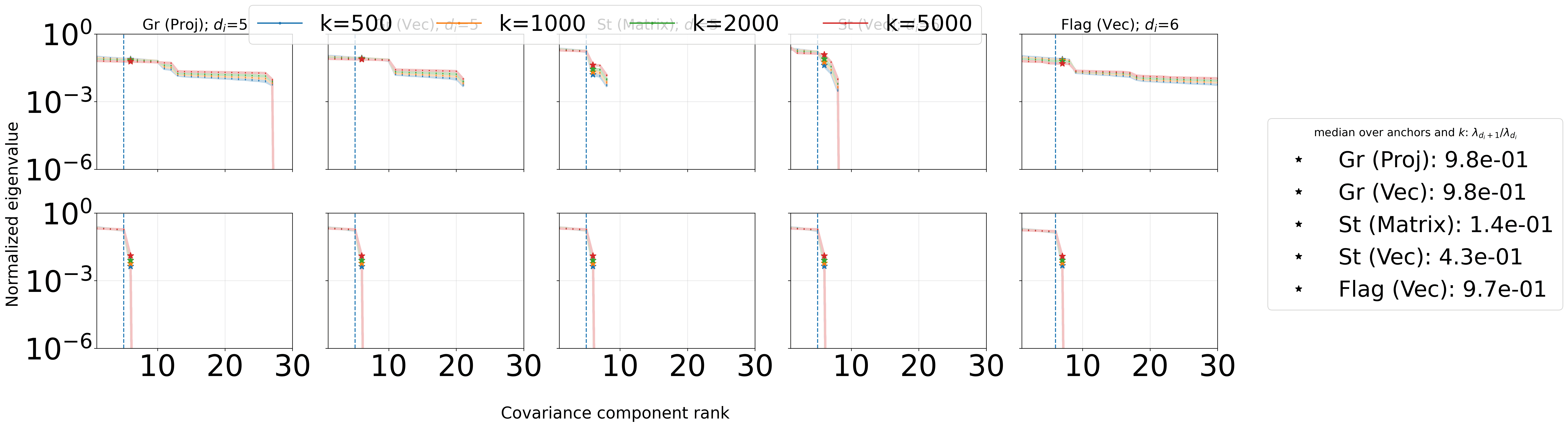}
        \caption{$N=10^5, d_i \sim 5$}
        \label{fig:local_covariance_spectra_N1000}
    \end{subfigure}
    \hfill
    \begin{subfigure}{0.49\textwidth}
        \centering
        \includegraphics[width=\textwidth]{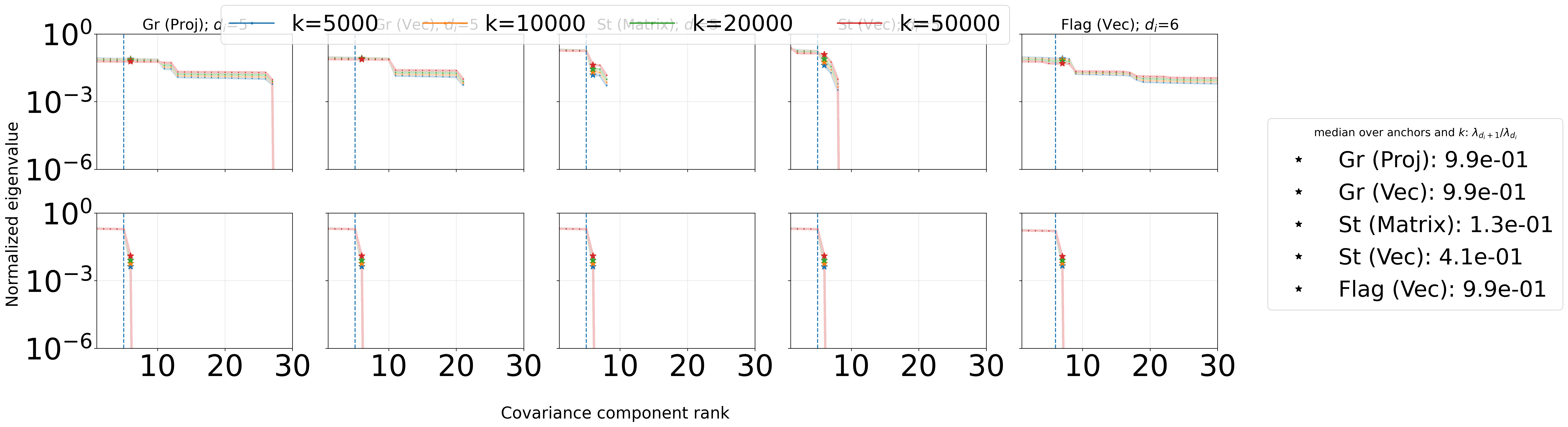}
        \caption{$N=10^6, d_i \sim 5$}
        \label{fig:local_covariance_spectra_N5000}
    \end{subfigure}

    \vspace{0.5em}

    \begin{subfigure}{0.49\textwidth}
        \centering
        \includegraphics[width=\textwidth]{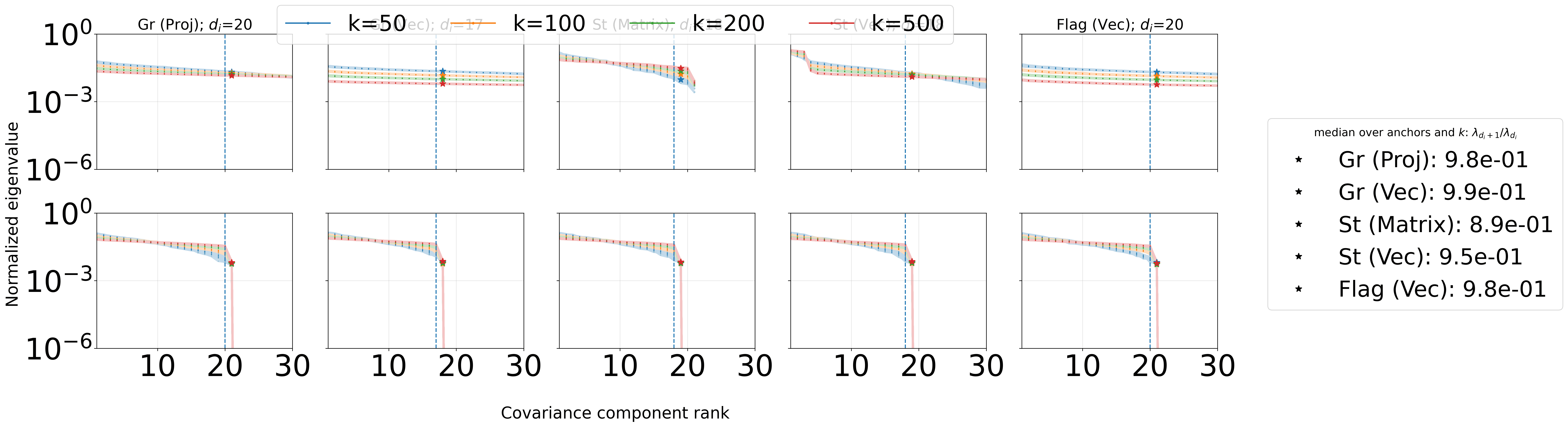}
        \caption{$N=10^4, d_i \sim 20$}
        \label{fig:local_covariance_spectra_N10000}
    \end{subfigure}
    \hfill
    \begin{subfigure}{0.49\textwidth}
        \centering
        \includegraphics[width=\textwidth]{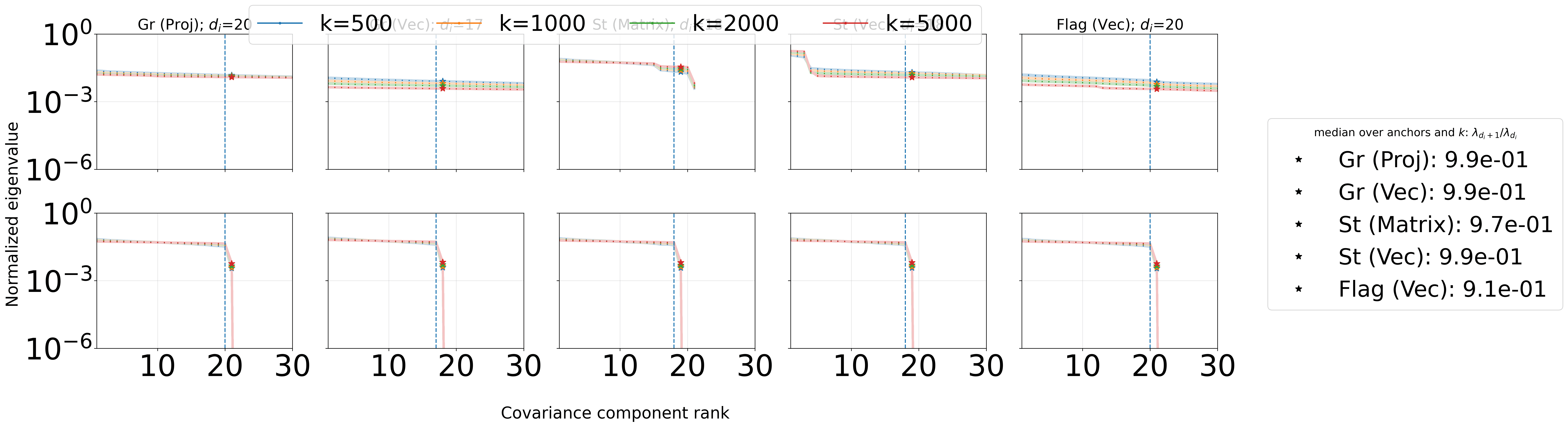}
        \caption{$N=10^5, d_i \sim 20$}
        \label{fig:local_covariance_spectra_diffd}
    \end{subfigure}

    \caption{
    Local covariance spectra from centered kNN neighborhoods. Each \quest manifold (top row) is compared to a matched sphere (bottom row) with the same $N$, $d_i$, and $d_a$. 
    \newtext{
    Matched spheres show an approximately flat tangent spectrum up to $d_i$, followed by a sharp drop. \quest embeddings show approximately flat staircases. However, unlike isotropic spheres, the tail possesses non-trivial variance. 
    This is shown by the stars in the figures showing the first drop in the spectra, that typically occurs around $d_i$, is typically $O(1)$.
    }
    }
    \label{fig:local_covariance_spectra}
\end{figure}

\newtext{
We numerically verify this in Figure~\ref{fig:local_covariance_spectra}. In this experiment, we compare \quest manifolds to matched spheres with the same intrinsic and ambient dimensions. Firstly, we observe approximate flat bands which confirm the symmetry-based arguments for our manifolds.
We observe approximately flat bands as expected, with the isotropic sphere exhibiting a strong degradation beyond its intrinsic dimension. 
In contrast, \quest manifolds  carry non-trivial signatures of their anisotropy in the spectral tail.
This provides an independent verification of our claim that \quest embeddings are anisotropic.
}

This diagnostic should be interpreted as an operational proxy, not as a complete differential-geometric invariant. It depends on $k$, $N$, the embedding, and the ambient metric, and it can mix tangent anisotropy, curvature, finite-sample effects, and density fluctuations. More precise notions could be based on the pullback metric, the second fundamental form, principal curvatures, or direction-dependent geodesic-to-Euclidean distortion, but estimating these robustly and scalably from high-dimensional samples is itself difficult. We therefore use local covariance anisotropy as a practical diagnostic aligned with the local kNN structure used by the IDE methods studied in this work.

\subsection{Comparison with other benchmarks}
\label{app:bm}

\begin{figure}
    \centering
    \includegraphics[width=0.99\linewidth]{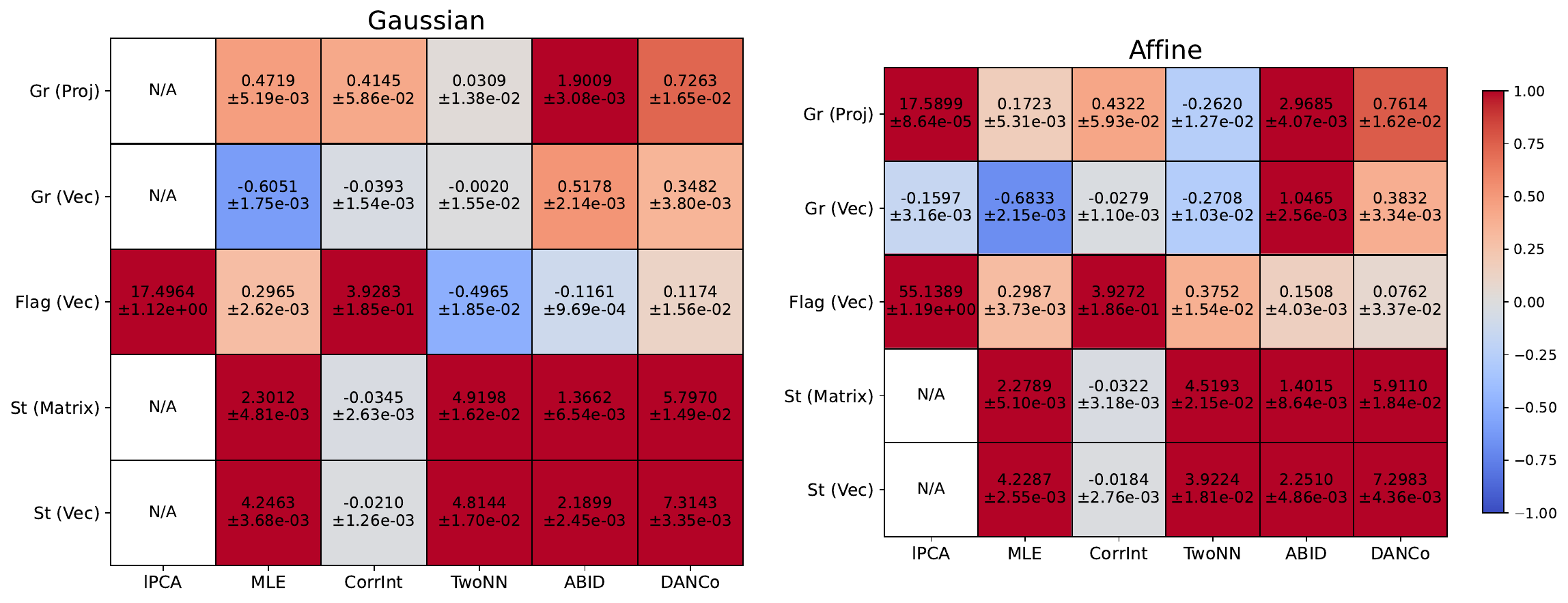}
    \caption{Relative performance of IDE methods on \quest\ manifolds compared against 
    (left) normal isotropic Gaussian vectors $\mathcal{N}$ and 
    (right) affine linear nullspace $\mathcal{L}$. 
    N/A indicates that the reference error is zero or numerically negligible, making the relative ratio unstable.}
    \label{fig:comp-rel-combined}
\end{figure}

Here we present results of comparing our manifolds against other standard manifolds. In Appx.~\ref{app:bm}, we present evidence that our manifolds are also adversarial as compared to other standard benchmarks. Note that one of our main considerations, was to compare these manifolds at fixed parameters and resources, hence we only include manifolds for which we can tune the $d_i$ and $d_a$. This is the reason why we exclude some manifolds like the Moebius strip, torus, Swiss rolls, etc.

\paragraph{Isotropic Gaussian vectors} Based on \citep{Hein2005Benchmark}, we compare our manifold against isotropic Gaussian vectors $\in \mR^{d_i}$ linearly embedded into $\mR^{d_a}$, cf. Fig.~\ref{fig:comp-rel-combined}

\paragraph{Affine spaces} Affine spaces are isomorphic to the linear nullspaces we considered in our main text. Given $d_i, d_a$, the nullspace of matrix $A \in \mR^{d_a \times d_a}$ with rank $d_a - d_i$ consists of a hyperplane of intrinsic dimension $d_i$. In order to preserve the "smoothness" of our manifold, we sample the coefficients of the basis vectors of the nullspace from the standard unit normal, cf. Fig.~\ref{fig:comp-rel-combined}.

For the purposes of this experiment, we run extensive scaling experiments by choosing small-dimensional manifolds. In particular we choose Grassmannians with ID of 2,3,4,5; Stiefels with ID of 3,5; Flags with ID of 4,12.%

\subsection{Scaling with sample size}
\label{app:scale-n}

The dependence of the intrinsic dimension estimate $\hat{d}_i$ comes from the fact that the fraction of neighbors in a small neighborhood for manifolds are given by \citep{Levina_MLE}

\begin{equation}
    \label{eq:sample-size}
    \frac{k}{N} = \Omega_{d_i} \rho(x) T_k^{d_i}
\end{equation}

\begin{figure}[h!]
    \centering
    \includegraphics[width=0.99\linewidth]{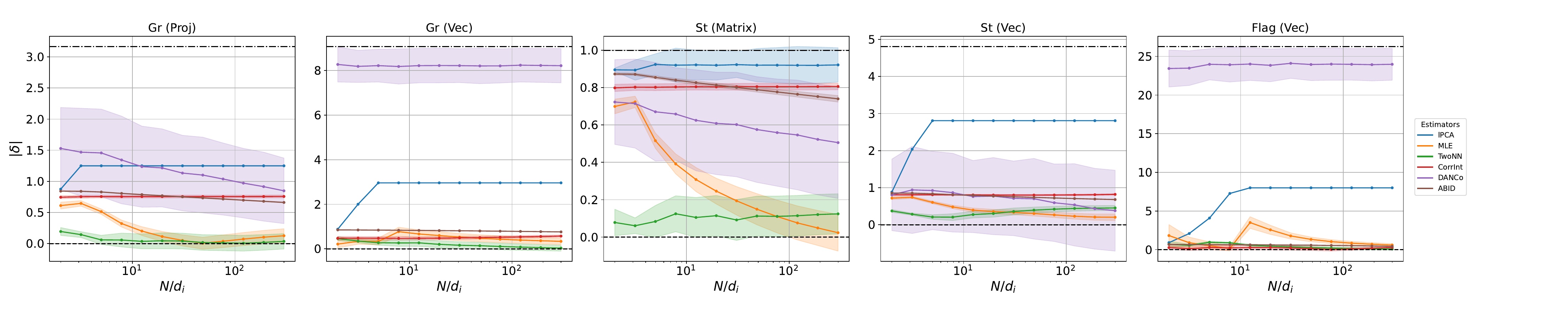}
    \caption{Performance of IDE methods as a function of $N/d_i$. We observe a gradual convergence with increasing sample size. The shaded region shows the 1-$\sigma$ error in the local ID estimates, averaged over three seeds.}
    \label{fig:n_di}
\end{figure}

Thus in order to get a uniform and locally dense sampling of points, we need to hold $k/(T_k^{d_i} N)$ fixed, which translates to the condition that $$ N \propto \exp(d_i)$$

We investigate the effect of changing $N$ while holding hyper-parameters fixed. We logarithmically sample $N$ so that $N/d_i$ goes from 2 to 300.  As Fig \ref{fig:n_di} suggests, the ID estimates saturates with exponentially many samples, thus we always choose the maximum value of $N$ possible for our experiments.

\subsection{Squeezing : comparison with isotropic manifolds}
\label{app:squeeze-compare-spheres}

\begin{figure}[t]
    \centering
    
    \begin{subfigure}{0.99\linewidth}
        \centering
        \includegraphics[width=\linewidth]{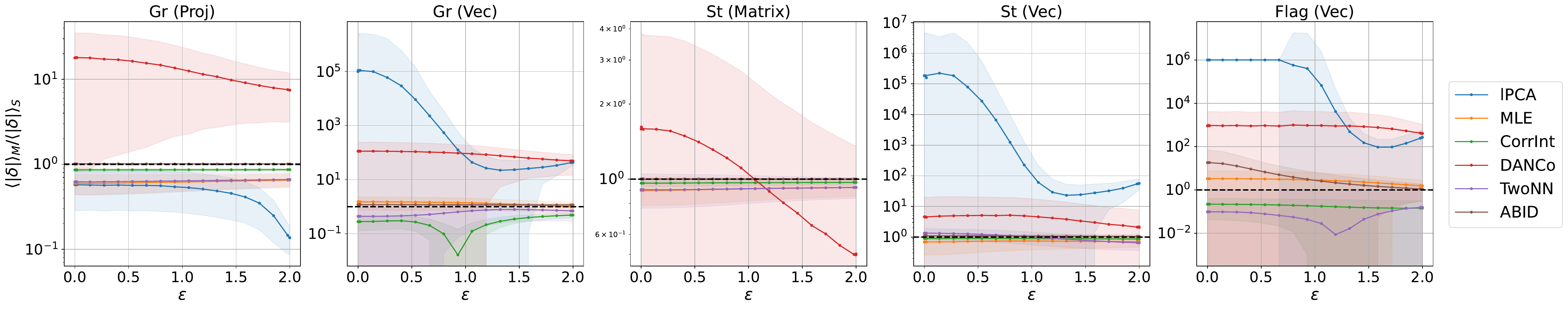}
        \caption{Hyperspheres}
        \label{fig:compare_distort_spheres_hyperspheres}
    \end{subfigure}
    
    \vspace{0.5em}
    
    \begin{subfigure}{0.99\linewidth}
        \centering
        \includegraphics[width=\linewidth]{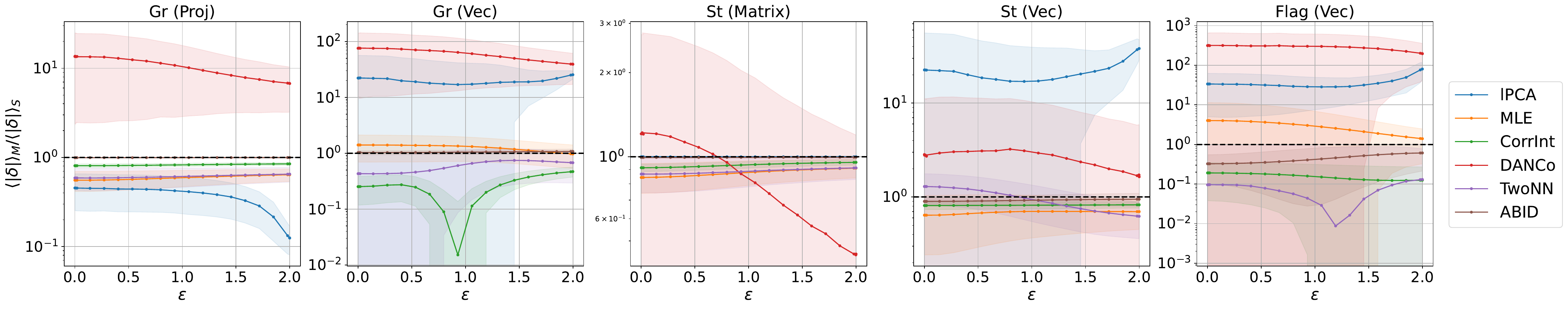}
        \caption{Linear nullspaces sampled from isotropic Gaussian}
        \label{fig:compare_distort_spheres_nullspaces}
    \end{subfigure}
    
    \caption{Comparison of errors when distorting \quest\ manifolds versus isotropic manifolds. The large change indicates that these manifolds become drastically more difficult to perform IDE on with increasing distortion, as opposed to \quest\ manifolds.}
    \label{fig:compare_distort_isotropic}
\end{figure}

We present in Fig.~\ref{fig:compare_distort_isotropic} the effect of distortion for spheres, as compared to \quest manifolds. In particular, we observe that spheres exhibit a very large increase in the error rate as anisotropy is increased. In conjunction with Section \ref{para:squeeze}, this demonstrates that \quest manifolds are already challenging enough for the IDE methods tested, and enhances it applicability as a more robust performance evaluator for IDE.

\subsection{Global estimators}
\label{app:global}
\begin{figure}[t!]
    \centering
    \includegraphics[width=0.98\linewidth]{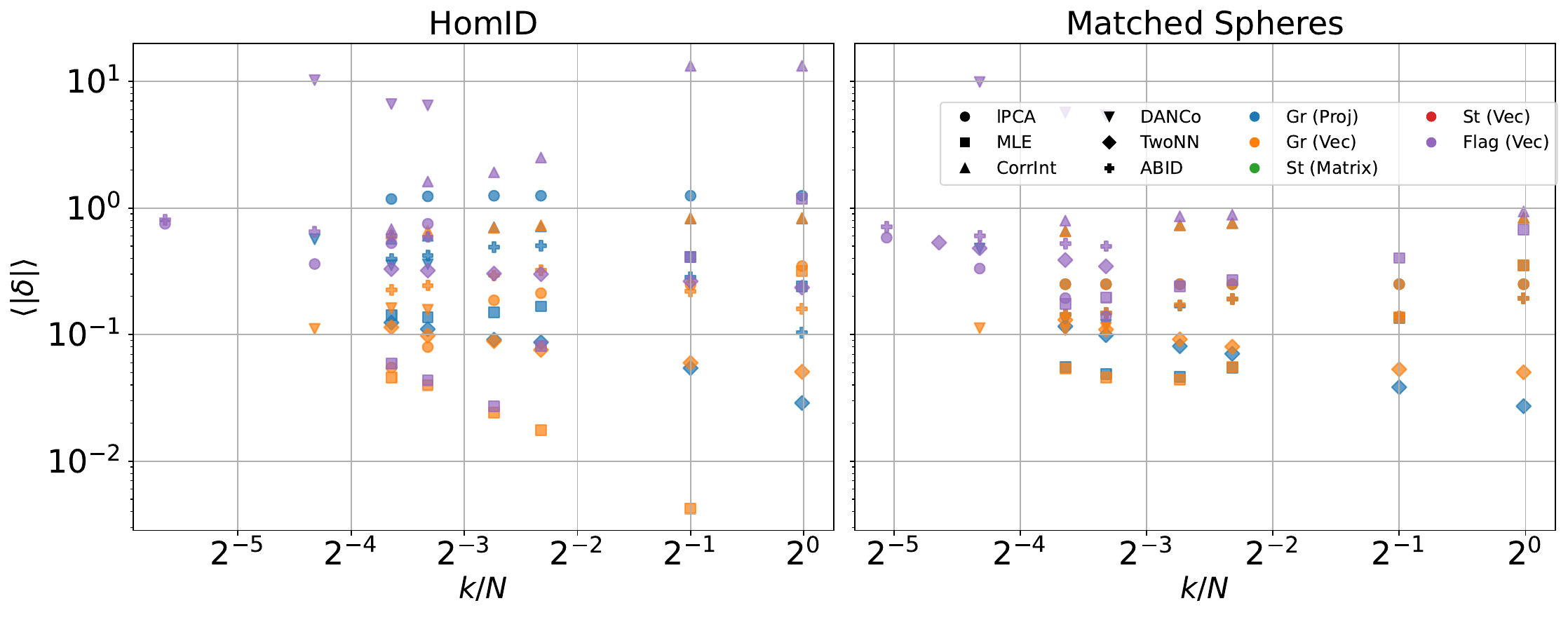}
    \caption{
    \newtext{
    Average estimation error as a function of the neighborhood ratio $k/N$. Most estimator--representation pairs remain approximately stable across the full range of $k/N$. A few exceptions are visible: MLE on Gr (Vec) exhibits a non-monotonic decrease followed by an increase as $k/N \to 1$, while DANCo produces comparatively large and noisy errors over much of the range. The origin of these estimator-specific behaviors is left to future work.
    }
    }
    \label{fig:local-to-global}
\end{figure}

\newtext{
Figure~\ref{fig:local-to-global} compares local estimates with the corresponding global estimates obtained as $k/N \to 1$. Overall, the estimators exhibit no consistent systematic dependence on $k/N$. 
We also list the comparison between the \quest manifolds and the corresponding matched spheres for $k = N$ for completion.  However, care must be taken to attribute this result to better performance, since the actual values of the error is still $\mathcal{O}(1)$ for most methods as shown in Figure~\ref{fig:local-to-global}.
Detailed investigation of this observation is left for future work.
Note that only lPCA and CorrInt transform exactly into their global estimator limits as $k \to N$. The other methods return a single global estimate of the ID.
}

\begin{table}[t]
\centering
\begin{tabular}{lcccccc}
\toprule
 & lPCA & MLE & CorrInt & DANCo & TwoNN & ABID \\
\midrule
Gr (Proj) & 4.000 & -0.319 & -0.014 & 2.012 & 0.060 & -0.465 \\
Gr (Vec) & 0.393 & -0.100 & -0.005 & 0.384 & 0.008 & -0.175 \\
St (Matrix) & -0.714 & 0.059 & -0.003 & 1.855 & -0.175 & -0.052 \\
St (Vec) & 0.014 & -0.063 & -0.004 & 2.648 & -0.178 & -0.117 \\
Flag (Vec) & 4.400 & 0.744 & 13.139 & 0.200 & -0.320 & 0.181 \\
\bottomrule
\end{tabular}
\caption{
\newtext{
$\langle|\delta|\rangle_{M}/\langle|\delta|\rangle_{S} - 1$ at the largest available $k/N$ (i.e.\ as $k/N \to 1$), for each HomID manifold $M$ and IDE method, relative to its matched-sphere baseline $S$. We notice that TwoNN leads to least deviation.
}
}
\label{tab:global_ratio}
\end{table}

\section{Additional Results}

\subsection{Signed v/s Unsigned error comparison}

We report the signed errors as in Table~\ref{tab:perf} to indicate if any methods consistently over- or under-estimate the ID. Table~\ref{tab:perf_signed} shows that all methods except for lPCA and DANCo under-estimate the intrinsic dimension.

\begin{table*}[t!]
\centering
\resizebox{0.99\textwidth}{!}{
\begin{tabular}{||l|ccccc||c||}
\toprule
\textbf{IDE Method} & \shortstack{Gr\\(Proj)} & \shortstack{Gr\\(Vec)}  & \shortstack{Flag\\(Vec)} & \shortstack{St\\(Matrix)} & \shortstack{St\\(Vec)} & \bf{\shortstack{Average\\(\quest)}} \\
\midrule
lPCA & $\textcolor{red}{0.1725\ \pm\ 0.1544}$ & $\textcolor{red}{0.0675\ \pm\ 0.0349}$ & $\textcolor{blue}{-0.1391\ \pm\ 0.0586}$ & $\textcolor{red}{0.0666\ \pm\ 0.1153}$ & $\textcolor{red}{0.0666\ \pm\ 0.1154}$ & $\textcolor{red}{0.0468\ \pm\ 0.1461}$ \\
MLE & $\textcolor{blue}{-0.0467\ \pm\ 0.0639}$ & $\textcolor{blue}{-0.1070\ \pm\ 0.0729}$ & $\textcolor{blue}{-0.2451\ \pm\ 0.0505}$ & $\textcolor{blue}{-0.1807\ \pm\ 0.0713}$ & $\textcolor{blue}{-0.1646\ \pm\ 0.0647}$ & $\textcolor{blue}{-0.1488\ \pm\ 0.0937}$ \\
CorrInt & $\textcolor{blue}{-0.7025\ \pm\ 0.0100}$ & $\textcolor{blue}{-0.7140\ \pm\ 0.0085}$ & $\textcolor{blue}{-0.8221\ \pm\ 0.0051}$ & $\textcolor{blue}{-0.7661\ \pm\ 0.0077}$ & $\textcolor{blue}{-0.7632\ \pm\ 0.0079}$ & $\textcolor{blue}{-0.7536\ \pm\ 0.0435}$ \\
TwoNN & $\textcolor{blue}{-0.0085\ \pm\ 0.0359}$ & $\textcolor{blue}{-0.0164\ \pm\ 0.0449}$ & $\textcolor{blue}{-0.1602\ \pm\ 0.0416}$ & $\textcolor{blue}{-0.1042\ \pm\ 0.0467}$ & $\textcolor{blue}{-0.1078\ \pm\ 0.0403}$ & $\textcolor{blue}{-0.0794\ \pm\ 0.0718}$ \\
ABID & $\textcolor{blue}{-0.0396\ \pm\ 0.1120}$ & $\textcolor{blue}{-0.0968\ \pm\ 0.0784}$ & $\textcolor{blue}{-0.3466\ \pm\ 0.0899}$ & $\textcolor{blue}{-0.1809\ \pm\ 0.0918}$ & $\textcolor{blue}{-0.1660\ \pm\ 0.0974}$ & $\textcolor{blue}{-0.1660\ \pm\ 0.1402}$ \\
DANCo & $\textcolor{red}{0.1194\ \pm\ 0.0433}$ & $\textcolor{red}{0.0301\ \pm\ 0.0534}$ & $\textcolor{red}{5.3542\ \pm\ 0.0712}$ & $\textcolor{blue}{-0.0171\ \pm\ 0.0626}$ & $\textcolor{red}{0.0103\ \pm\ 0.0569}$ & $\textcolor{red}{1.0994\ \pm\ 2.1287}$ \\
\midrule
\textbf{Average} & $\textcolor{blue}{-0.0842\ \pm\ 0.3007}$ & $\textcolor{blue}{-0.1394\ \pm\ 0.2700}$ & $\textcolor{red}{0.6068\ \pm\ 2.1362}$ & $\textcolor{blue}{-0.1971\ \pm\ 0.2792}$ & $\textcolor{blue}{-0.1875\ \pm\ 0.2812}$ & $\textcolor{blue}{-0.0003\ \pm\ 1.0346}$ \\
\bottomrule
\end{tabular}%
}
\vspace{0.08in}
\caption{Mean signed relative error $\delta$ (mean $\pm$ standard deviation) for various methods on \quest\ manifolds. Positive values indicate overestimation of the intrinsic dimension, while negative values indicate underestimation. Most methods exhibit a consistent under-estimation bias across all manifolds, with lPCA and DANCo over-estimating consistently on certain manifold combinations.}
\label{tab:perf_signed}
\end{table*}

\subsection{Additional plots for anisotropic distortion and noise perturbation}
\label{app:add-squeeze-noise}

We present here the additional plots for anisotropic distortion and additive noise perturbation for completeness. The plots presented in the main text are a subset of these plots.

\begin{figure}[h!]
    \centering

    \begin{subfigure}[t]{0.99\linewidth}
        \centering
        \includegraphics[width=\linewidth]{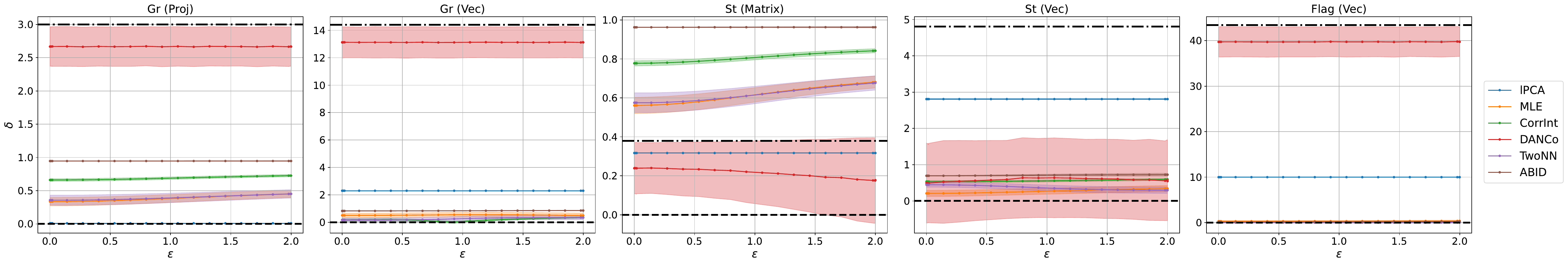}
        \caption{Anisotropic distortion experiments (Sec.~\ref{para:squeeze}).}
        \label{fig:squeeze-full}
    \end{subfigure}

    \vspace{0.5em}

    \begin{subfigure}[t]{0.99\linewidth}
        \centering
        \includegraphics[width=\linewidth]{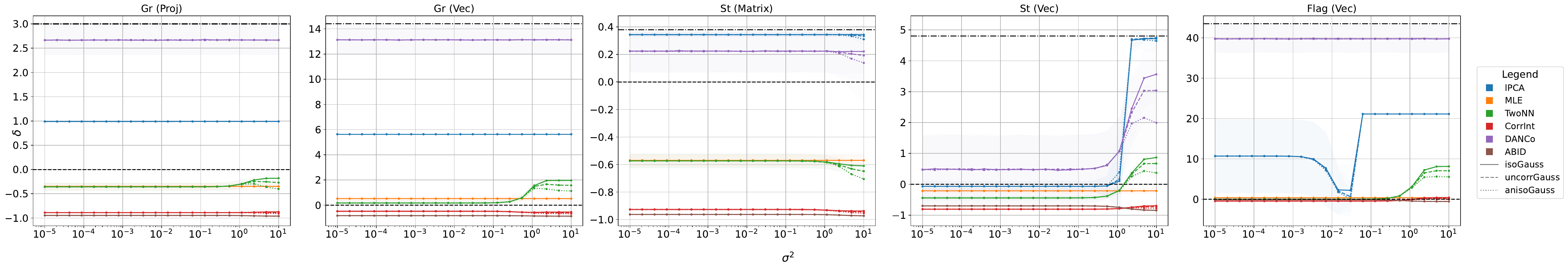}
        \caption{Additive perturbation experiments (Sec.~\ref{para:noise}).}
        \label{fig:noise-full}
    \end{subfigure}

    \caption{Full experimental results on all \quest\ manifolds.
    Top: anisotropic distortion (squeeze) experiments. 
    Bottom: additive perturbation (noise) experiments.}

    \label{fig:full-experiments}
\end{figure}

\subsection{Null results on correlation of IDE performance}
\label{app:null}

We analyze IDE performance in terms of different statistical and geometric features of the data.

\paragraph{Statistical properties } We look at three important features of the data covariance matrix $\mathbf{\Sigma}$: (1) the total variance given by Tr $\mathbf{\Sigma}$, (2) the variance dispersion index (VDI) \citep{Watanabe2022_VDI} which is a direct measure of the global anisotropy of the data and is given by $\frac{\text{Var}(\lambda)}{(\text{Mean}(\lambda))^2}$ ( $\lambda$ refers to the eigenvalues of $\mathbf{\Sigma}$), and (3) the $\langle R \rangle^2$ value defined as $\frac{1}{d_a^2} \sum_{i,j} \frac{\mathbf{\Sigma}_{ij}^2}{\mathbf{\Sigma}_{ii} \mathbf{\Sigma}_{jj}}$, which captures the inter-component correlation. Note that anisotropy does not imply that components are correlated, but correlated components necessarily imply anisotropy.

We observe very little statistically significant correlation of IDE performance with the total variance and $\langle R^2 \rangle$ for these methods. There is a slight negative correlation of performance with VDI, the effect being most prominent for the angle-based methods DANCo and ABID, but nothing statistically significant.

\begin{figure}[h!]
    \centering
    \begin{subfigure}{0.99\linewidth}
        \centering
        \includegraphics[width=\linewidth]{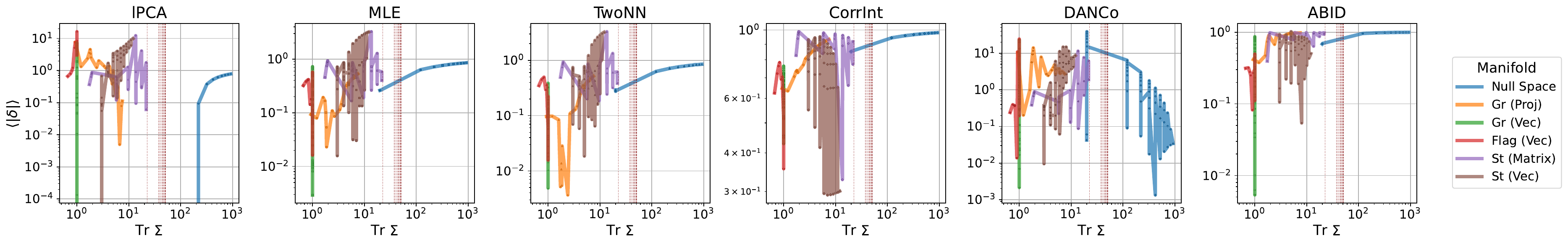}
        \caption{Tr $\Sigma$}
        \label{fig:dc-tr}
    \end{subfigure}
    \vspace{0.5cm}
    \begin{subfigure}{0.99\linewidth}
        \centering
        \includegraphics[width=\linewidth]{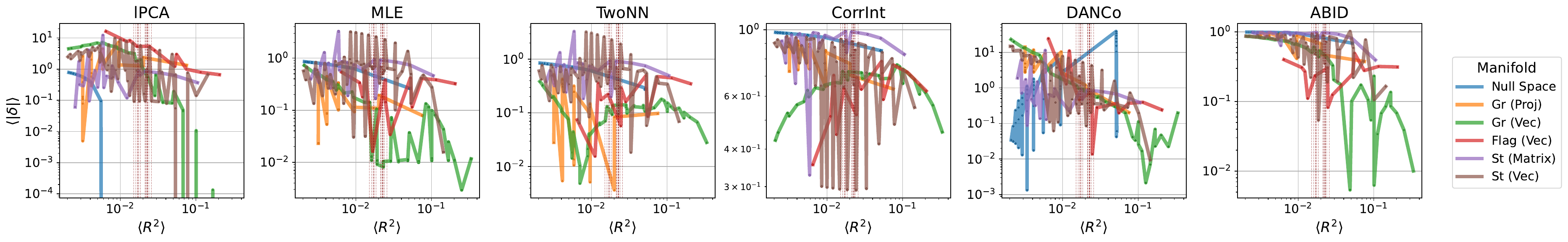}
        \caption{$\langle R^2 \rangle$}
        \label{fig:dc-rsq}
    \end{subfigure}
    \begin{subfigure}{0.99\linewidth}
        \centering
        \includegraphics[width=\linewidth]{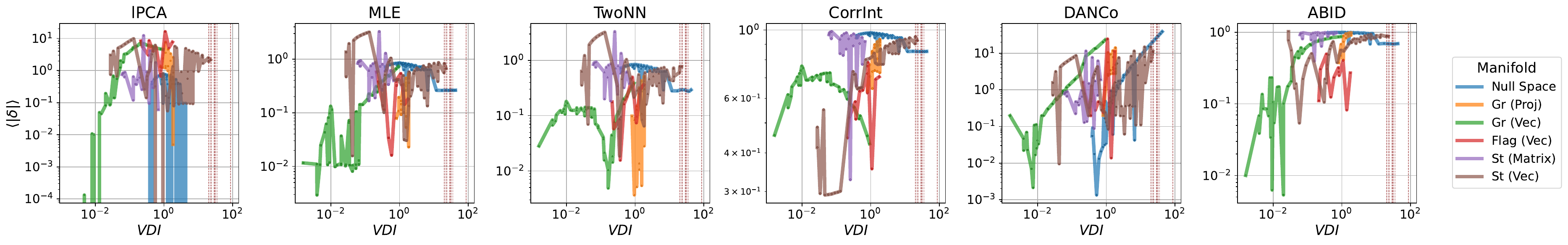}
        \caption{$VDI$}
        \label{fig:dc-vdi}
    \end{subfigure}

    \caption{IDE performance seems to be almost independent for these various quantities of interest. The light vertical maroon lines represent the corresponding quantities for different classes in MNIST. }
    \label{fig:combined-data-cov}
\end{figure}

The plots for these results are presented in Fig.~\ref{fig:combined-data-cov}.

\paragraph{Geometric Properties}
\label{app:geom}

We measure the local mean curvature $H(x)$ and the local density $\rho(x)$. From these two measurements, we calculate a dimensionless parameter $\kappa(x) = \rho/H^{d_i}$. In order to find the curvature, we fit a quadratic surface in a local neighborhood and use standard results due to Gauss. We sample a kNN-neighborhood and estimate the density based on the ratio $k/\Omega_{d_i} T_k^{d_i}$. We sample 5 logarithmically-spaced $k$ values and choose the median value as the desired geometric property. However, we fail to observe any definite correlation of IDE performance with these geometrical quantities as well. The tabulated results are shown in Table~\ref{tab:avg_values}.

\begin{table}[ht]
\centering
\small

\begin{tabular}{l|ccccc}
\hline
Quantity            & Gr (Vec) & Gr (Proj) & St (Matrix) & St (Vec) & Flag (Vec) \\
\hline
\hline
$\langle H \rangle$ &
$1.0313 \pm 0.0313$ &
$1.1417 \pm 0.0525$ &
$0.6702 \pm 0.0431$ &
$0.4667 \pm 0.2115$ &
$0.5348 \pm 0.1300$ \\
$\langle \rho \rangle$ &
$0.8527 \pm 0.1396$ &
$0.5933 \pm 0.1830$ &
$0.1766 \pm 0.1081$ &
$0.0373 \pm 0.0153$ &
$0.2846 \pm 0.1263$ \\
$\langle \kappa \equiv \rho/H^{d_i} \rangle$ &
$0.7758 \pm 0.2164$ &
$0.3736 \pm 0.0821$ &
$0.9097 \pm 0.6318$ &
$0.8430 \pm 0.4767$ &
$2.1033(*) \pm 2.1022$ \\

\hline
$\langle|\delta|\rangle$ & 0.2023 $\pm$ 0.0030 & 0.5785 $\pm$ 0.0045 & 0.5793 $\pm$ 0.0040 & 0.8502 $\pm$ 0.0024 & 2.4085 $\pm$ 0.0700 \\
\hline
\end{tabular}

\captionsetup{justification=centering}
\caption{Average absolute values by manifold (manifolds sorted by increasing $\langle|\delta|\rangle$).\\(*) This number was skewed by a manifold with intrinsic dimension $d_i = 12$, which we do not include in the table. Including that manifold gives 4080.9902.}
\label{tab:avg_values}
\end{table}

\subsection{Additional plots on real datasets}
\label{app:mnist}

\begin{figure}[h]
    \centering

    \begin{subfigure}{0.9\linewidth}
        \centering
        \includegraphics[width=\linewidth]{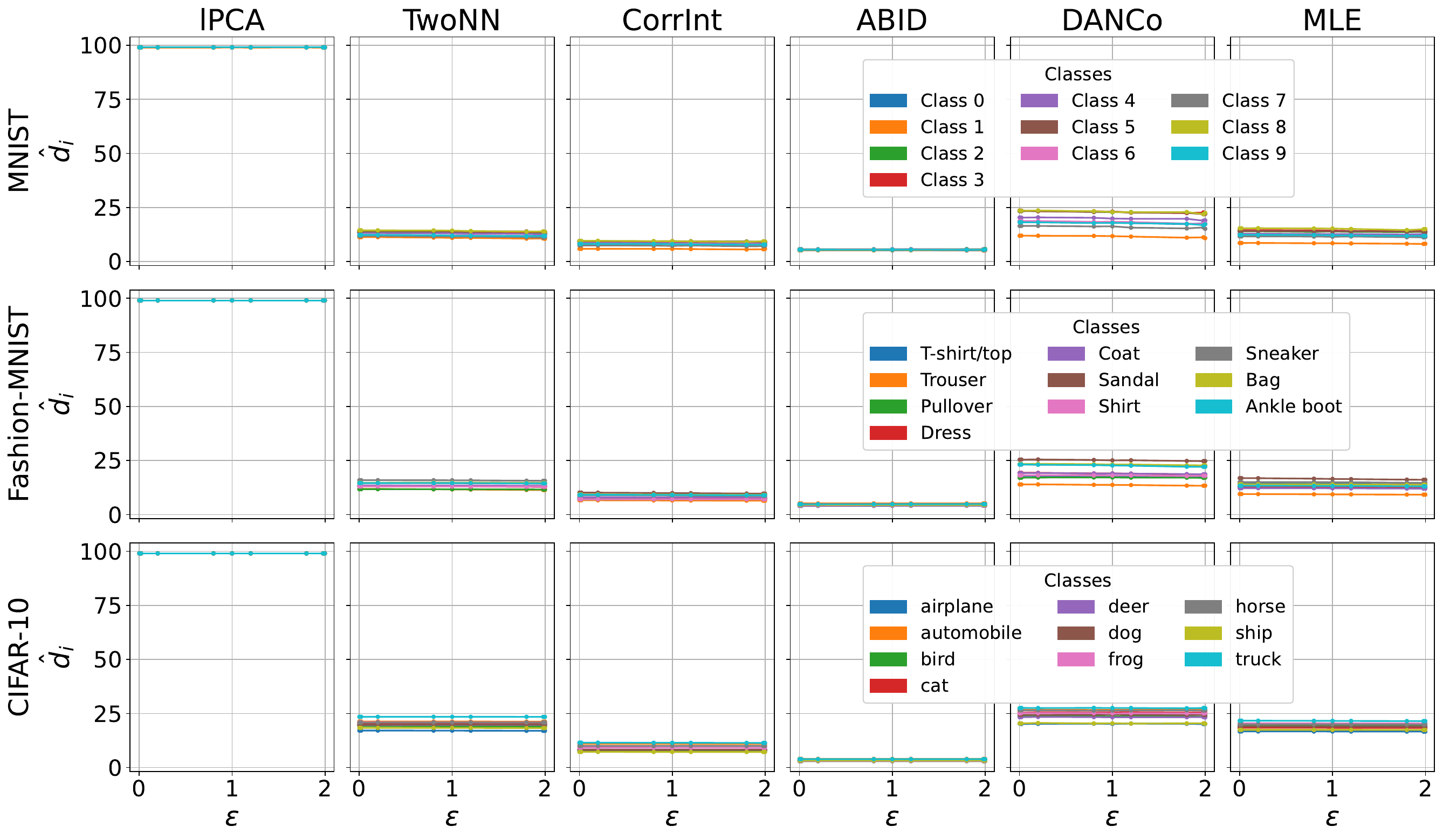}
        \caption{Squeezing perturbation}
        \label{fig:squeeze-mnist}
    \end{subfigure}

    \vspace{0.5cm}

    \begin{subfigure}{0.9\linewidth}
        \centering
        \includegraphics[width=\linewidth]{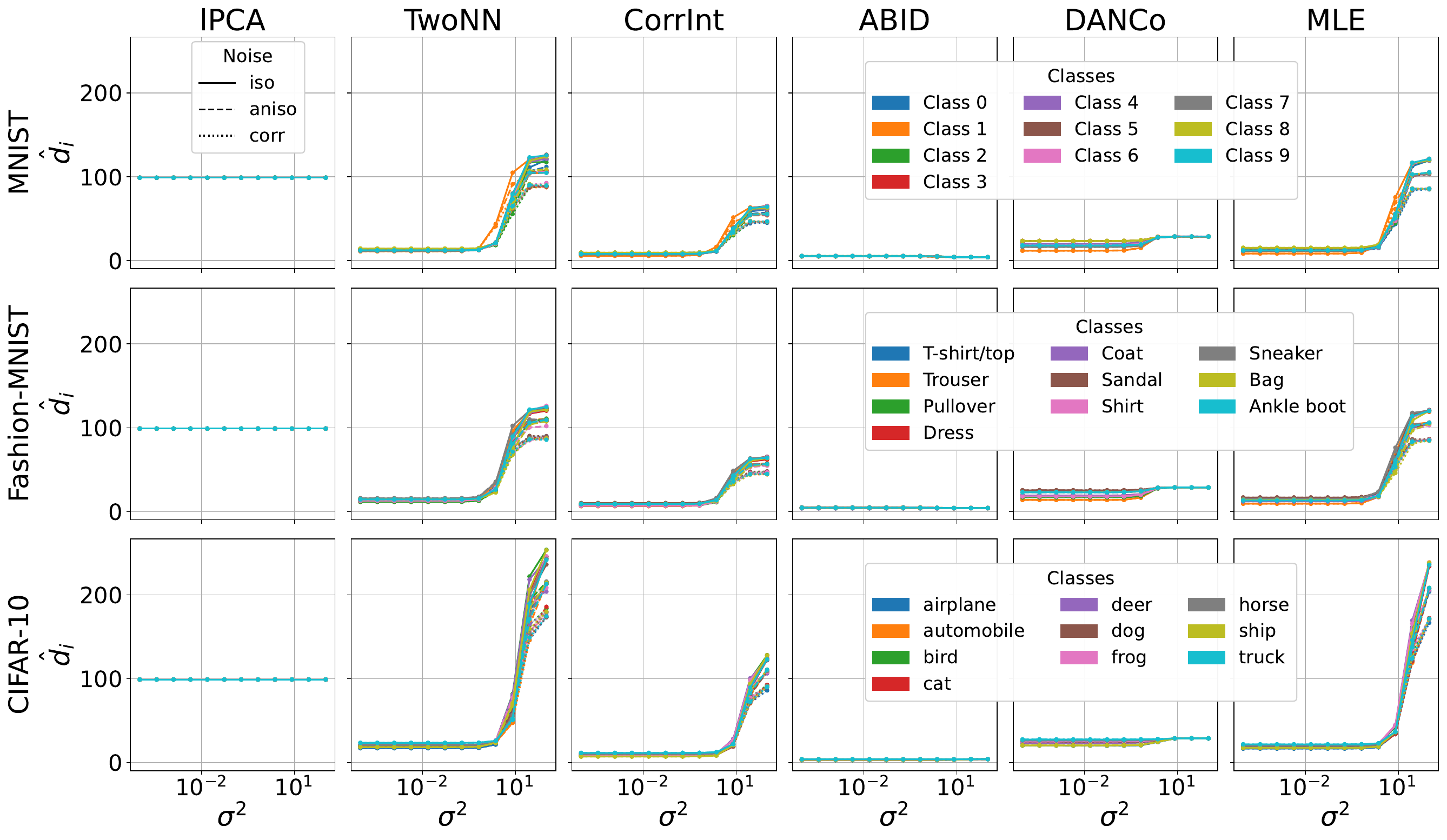}
        \caption{Additive noise perturbation}
        \label{fig:add-noise-mnist}
    \end{subfigure}

    \caption{Effect of (a) distortion and (b) additive noise perturbation on MNIST, FashionMNIST and CIFAR-10 datasets. We observe minimal degradation in (a) and degradation when $\sigma^2 \sim \mO(1)$ in (b), both observed for \quest manifolds as well.}
    \label{fig:mnist-full}
\end{figure}

We provide additional plots here regarding our distortion experiments in MNIST. As emphasized in the main text, our observations are similar qualitatively to the analogous experiments for \quest manifolds. See Fig.~\ref{fig:mnist-full}

\subsection{Additional lPCA interpretation plots}
\label{app:lpca}

\begin{figure}[ht!]
    \centering
    \includegraphics[width=0.98\linewidth]{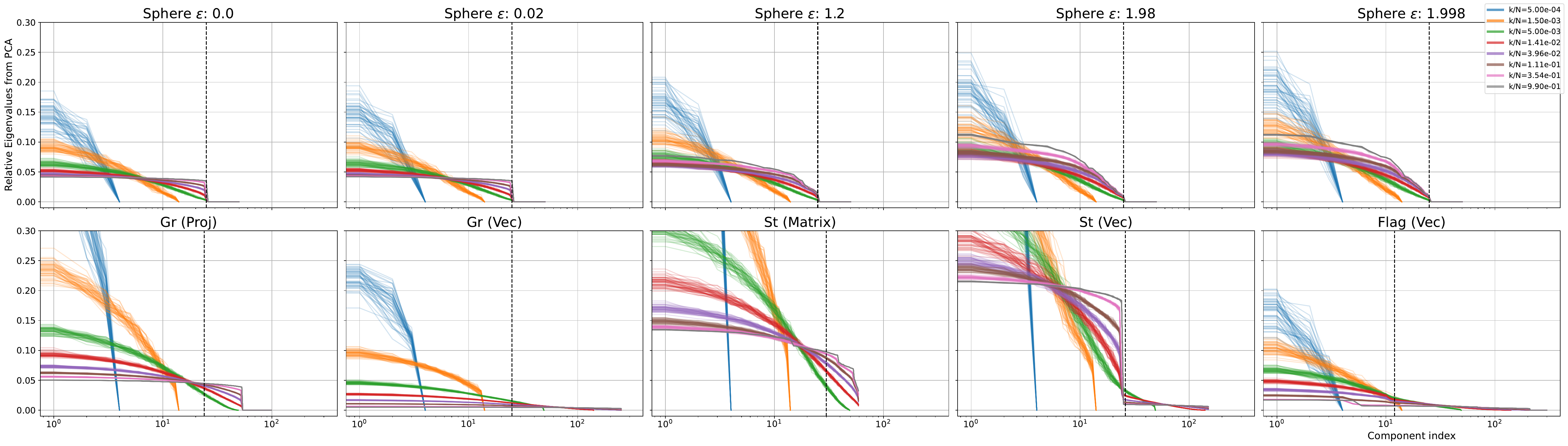}
    \caption{Extra lPCA plots for various manifolds. The parameter $\epsilon$ denotes the degree of anisotropy, with larger $\epsilon$ indicating greater anisotropy. Only the undistorted sphere (top left) shows the expected sharp drop at the ID.}
    \label{fig:lpca-interp-app}
\end{figure}

We present additional plots on lPCA here in the appendix. As alluded to in the main text, our main conjecture is that the kNN-based IDE methods fail due to deviations of their expected distributions. In the case of lPCA, this is true for distorted spheres and \quest manifolds, cf. Fig.~\ref{fig:lpca-interp-app}.

\subsection{Mechanistic failure analysis extended to other IDEs}
\label{app:mech-interp}

\newtext{
We present a summary of the distributional shifts that occur in \quest embeddings, comparing them to distorted and undistorted spheres.
}

\newtext{
For MLE and TwoNN, we observe shifts in local distance-ratio statistics. CorrInt shows deviations in the local growth rate. DANCo shows two distinct failures. Its distance marginal often concentrates at large normalized radii with small variance. Its angular marginal shows positive cosine bias, which means local neighbor directions are too aligned. Distorted spheres have a very strong angular collapse. \quest embeddings show milder but structured angular bias, with Gr (Vec) and Flag (Vec) most separated from the undistorted sphere. Table~\ref{tab:mech-interp} summarizes these distributional shifts using moments, Wasserstein distances, and Kolmogorov-Smirnov distances.
}

\begin{table}[t!]
\centering
\scriptsize
\setlength{\tabcolsep}{2pt}
\resizebox{\textwidth}{!}{%
\begin{tabular}{lcccccccc}
\toprule
Method & $M_1^0/M_1^d$ & $M_2^0/M_2^d$ & $M_3^0/M_3^d$ & $M_4^0/M_4^d$ & $M_5^0/M_5^d$ & $M_6^0/M_6^d$ & KS$_0$/KS$_d$ & W$_0$/W$_d$ \\
\midrule
\multicolumn{9}{l}{Gr (Proj) $(d_i,d_a)=(24,100)$} \\
MLE & 0.63/0.73 & 0.42/0.16 & \textcolor{red}{0.30/0.03} & \textcolor{red}{0.25/5.87e-03} & \textcolor{red}{0.23/1.60e-03} & \textcolor{red}{0.23/5.31e-04} & 0.19/0.10 & 0.56/0.57 \\
TwoNN & 0.68/0.86 & 0.49/0.63 & 0.37/0.42 & 0.29/0.26 & 0.22/0.15 & 0.16/0.08 & \textcolor{red}{0.17/0.04} & 0.45/0.17 \\
CorrInt & 1.39/0.77 & \textcolor{red}{1.84/0.55} & \textcolor{red}{2.27/0.38} & \textcolor{red}{2.65/0.26} & \textcolor{red}{3.00/0.17} & \textcolor{red}{3.33/0.11} & 0.55/0.30 & 0.30/0.34 \\
DANCo dist. & 1.08/1.10 & 1.17/1.17 & 1.26/1.21 & 1.35/1.24 & 1.45/1.26 & 1.55/1.27 & 0.52/0.31 & 0.06/0.08 \\
DANCo angle & 1.21/0.42 & \textcolor{red}{1.25/0.18} & \textcolor{red}{1.23/0.08} & \textcolor{red}{1.16/0.04} & \textcolor{red}{1.07/0.02} & \textcolor{red}{0.97/8.95e-03} & \textcolor{red}{0.26/0.93} & \textcolor{red}{0.07/0.49} \\
\midrule
\multicolumn{9}{l}{Gr (Vec) $(d_i,d_a)=(25,252)$} \\
MLE & 0.44/0.77 & 0.21/0.55 & \textcolor{red}{0.11/0.39} & \textcolor{red}{0.05/0.28} & \textcolor{red}{0.03/0.21} & \textcolor{red}{0.01/0.17} & \textcolor{red}{0.32/0.09} & \textcolor{red}{0.86/0.20} \\
TwoNN & 0.62/0.78 & 0.41/0.52 & 0.28/0.30 & 0.20/0.15 & 0.15/0.07 & \textcolor{red}{0.12/0.03} & 0.20/0.07 & 0.54/0.25 \\
CorrInt & 2.35/1.57 & 6.11/2.76 & \textcolor{red}{17.10/5.28} & \textcolor{red}{50.23/10.69} & \textcolor{red}{$1.52{\times}10^{2}$/22.46} & \textcolor{red}{$4.70{\times}10^{2}$/48.16} & 0.75/0.55 & 1.02/0.65 \\
DANCo dist. & 1.12/1.14 & 1.25/1.25 & 1.39/1.34 & 1.55/1.42 & 1.71/1.49 & 1.89/1.55 & 0.72/0.42 & 0.09/0.11 \\
DANCo angle & 1.42/0.49 & \textcolor{red}{1.67/0.24} & \textcolor{red}{1.78/0.12} & \textcolor{red}{1.76/0.06} & \textcolor{red}{1.66/0.03} & \textcolor{red}{1.49/0.01} & 0.60/0.95 & \textcolor{red}{0.13/0.44} \\
\midrule
\multicolumn{9}{l}{St (Matrix) $(d_i,d_a)=(30,60)$} \\
MLE & 0.72/0.53 & \textcolor{red}{0.53/0.04} & \textcolor{red}{0.39/2.41e-03} & \textcolor{red}{0.29/1.53e-04} & \textcolor{red}{0.21/1.10e-05} & \textcolor{red}{0.14/8.67e-07} & 0.14/0.11 & 0.46/1.28 \\
TwoNN & 0.73/0.90 & 0.53/0.48 & 0.37/0.13 & \textcolor{red}{0.25/0.03} & \textcolor{red}{0.15/5.88e-03} & \textcolor{red}{0.09/1.37e-03} & \textcolor{red}{0.13/0.04} & 0.40/0.18 \\
CorrInt & 1.41/0.79 & \textcolor{red}{1.88/0.52} & \textcolor{red}{2.38/0.29} & \textcolor{red}{2.88/0.14} & \textcolor{red}{3.38/0.06} & \textcolor{red}{3.91/0.02} & 0.55/0.35 & 0.29/0.33 \\
DANCo dist. & 1.06/1.09 & 1.12/1.15 & 1.19/1.18 & 1.25/1.20 & 1.32/1.22 & 1.39/1.22 & 0.47/0.30 & 0.05/0.08 \\
DANCo angle & 1.16/0.44 & \textcolor{red}{1.22/0.19} & \textcolor{red}{1.23/0.09} & \textcolor{red}{1.21/0.04} & \textcolor{red}{1.16/0.02} & \textcolor{red}{1.10/9.40e-03} & \textcolor{red}{0.22/0.93} & \textcolor{red}{0.05/0.49} \\
\midrule
\multicolumn{9}{l}{St (Vec) $(d_i,d_a)=(26,151)$} \\
MLE & 0.91/0.69 & \textcolor{red}{0.83/0.25} & \textcolor{red}{0.81/0.07} & \textcolor{red}{0.85/0.02} & \textcolor{red}{0.95/9.19e-03} & \textcolor{red}{1.09/3.89e-03} & 0.05/0.09 & \textcolor{red}{0.14/0.77} \\
TwoNN & 1.15/1.48 & 1.43/1.52 & 1.70/0.95 & \textcolor{red}{1.85/0.48} & \textcolor{red}{1.83/0.24} & \textcolor{red}{1.71/0.13} & 0.07/0.17 & 0.24/0.67 \\
CorrInt & 1.43/0.73 & \textcolor{red}{2.20/0.50} & \textcolor{red}{3.63/0.32} & \textcolor{red}{6.29/0.20} & \textcolor{red}{11.21/0.12} & \textcolor{red}{20.32/0.07} & 0.40/0.50 & 0.28/0.36 \\
DANCo dist. & 1.01/1.03 & 1.02/1.02 & 1.04/0.99 & 1.06/0.95 & 1.08/0.91 & 1.10/0.87 & 0.15/0.25 & \textcolor{red}{0.02/0.07} \\
DANCo angle & 1.12/0.41 & \textcolor{red}{1.15/0.18} & \textcolor{red}{1.15/0.08} & \textcolor{red}{1.14/0.04} & \textcolor{red}{1.12/0.02} & \textcolor{red}{1.10/0.01} & \textcolor{red}{0.14/0.91} & \textcolor{red}{0.04/0.50} \\
\midrule
\multicolumn{9}{l}{Flag (Vec) $(d_i,d_a)=(11,300)$} \\
MLE & 0.47/0.64 & 0.28/0.41 & 0.20/0.26 & 0.15/0.17 & 0.11/0.12 & 0.07/0.08 & 0.32/0.15 & 0.64/0.32 \\
TwoNN & 0.79/1.00 & 0.70/1.04 & 0.68/1.10 & 0.69/1.16 & 0.71/1.23 & 0.71/1.32 & \textcolor{red}{0.12/0.03} & \textcolor{red}{0.25/0.03} \\
CorrInt & 1.46/1.05 & 1.93/0.94 & \textcolor{red}{2.39/0.70} & \textcolor{red}{2.83/0.45} & \textcolor{red}{3.29/0.26} & \textcolor{red}{3.78/0.15} & 0.60/0.25 & 0.48/0.27 \\
DANCo dist. & 1.17/1.12 & 1.37/1.23 & 1.59/1.33 & 1.84/1.42 & 2.12/1.50 & 2.44/1.59 & 0.56/0.31 & 0.11/0.08 \\
DANCo angle & \textcolor{red}{1.92/0.57} & \textcolor{red}{1.66/0.30} & \textcolor{red}{1.61/0.16} & \textcolor{red}{1.28/0.08} & \textcolor{red}{1.05/0.04} & \textcolor{red}{0.82/0.02} & 0.58/0.74 & 0.21/0.32 \\
\bottomrule
\end{tabular}%
}
\caption{Raw-moment and two-sample summaries for one sample from each HomID manifold. Red entries mark ratios where one value is at least three times the other. For $k=1,\ldots,6$, $M_k^0=\mathbb{E}_M[X^k]/\mathbb{E}_{S_0}[X^k]$ and $M_k^d=\mathbb{E}_M[X^k]/\mathbb{E}_{S_d}[X^k]$, where $S_0$ is the matched undistorted sphere and $S_d$ is the matched distorted sphere at $\epsilon=1.998$. W and KS denote the Wasserstein distance and Kolmogorov-Smirnov distance respectively.}
\label{tab:mech-interp}
\end{table}

\end{document}